\pdfoutput=1

\documentclass[pdflatex,sn-chicago]{sn-jnl}
\usepackage{times}
\usepackage{latexsym}
\usepackage[T1]{fontenc}
\usepackage[utf8]{inputenc}
\usepackage{microtype}
\usepackage{inconsolata}
\usepackage{graphicx}
\usepackage{xcolor}
\usepackage{booktabs}
\usepackage{subcaption}
\usepackage{amsmath}
\usepackage{cleveref}
\usepackage{tabularx}
\usepackage{color}
\usepackage{xcolor}
\usepackage{placeins}
\usepackage{array}
\usepackage{xparse}

\newcolumntype{L}[1]{>{\raggedright\let\newline\\\arraybackslash\hspace{0pt}}p{#1}}
\newcolumntype{C}[1]{>{\centering\let\newline\\\arraybackslash\hspace{0pt}}p{#1}}
\newcolumntype{R}[1]{>{\raggedleft\let\newline\\\arraybackslash\hspace{0pt}}p{#1}}
\newcommand{\casestudybox}[1]{\begin{center}\begin{minipage}[t]{\textwidth}\fcolorbox{gray}{gray!6}{\begin{minipage}[t]{\dimexpr\textwidth-2\fboxsep-2\fboxrule\relax} #1 \end{minipage}}\end{minipage}\end{center}}
\newcommand{\RNum}[1]{\uppercase\expandafter{\romannumeral #1\relax}}
\definecolor{c_255_197_197}{RGB}{255,197,197}
\definecolor{c_255_74_74}{RGB}{255,74,74}
\newcommand{\namewithnumber}[2]{{#1 \par \tiny{(#2)}}}
\NewDocumentCommand\judgeemoji{}{\raisebox{-0.06cm}{\includegraphics[scale=0.14]{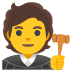}}}
\NewDocumentCommand\computeremoji{}{\raisebox{-0.06cm}{\includegraphics[scale=0.14]{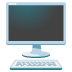}}}

\jyear{2023}%

\begin{document}

\title[Challenges in Explanation Quality Evaluation]{Challenges in Explanation Quality Evaluation}

\author*[1,2]{\fnm{Hendrik} \sur{Schuff}}\email{\color{black}Hendrik.Schuff@de.bosch.com}
\author[1]{\fnm{Heike} \sur{Adel}}\email{\color{black}Heike.Adel@de.bosch.com}
\author[3]{\fnm{Peng} \sur{Qi}}\email{\color{black}PengQi@cs.stanford.edu}
\author[2]{\fnm{Ngoc Thang} \sur{Vu}}\email{\color{black}Thang.Vu@ims.uni-stuttgart.de}

\affil[1]{\orgname{Bosch Center for Artificial Intelligence}, \city{Renningen}, \country{Germany}}
\affil[2]{\orgdiv{IMS}, \orgname{University of Stuttgart}, \city{Stuttgart}, \country{Germany}}
\affil[3]{\orgname{AWS AI Labs}, \city{Seattle}, \country{USA}. Work done prior to joining Amazon}

\abstract{
While much research focused on producing explanations, it is still unclear how the produced explanations' quality can be evaluated in a meaningful way.
Today's predominant approach is to quantify explanations using proxy scores which compare explanations to (human-annotated) gold explanations.
This approach assumes that explanations which reach higher proxy scores will also provide a greater benefit to human users.
In this paper, we present problems of this approach.
Concretely, we (i) formulate desired characteristics of explanation quality, (ii) describe how current evaluation practices violate them, and (iii) support our argumentation with initial evidence from a crowdsourcing case study in which we investigate the explanation quality of state-of-the-art explainable question answering systems. 
We find that proxy scores correlate poorly with human quality ratings and, additionally, become less expressive the more often they are used (i.e. following Goodhart's law).
Finally, we propose guidelines to enable a meaningful evaluation of explanations to drive the development of systems that provide tangible benefits to human users.
}



\maketitle

\newpage
\section{Introduction}
\begin{figure}
\centering
\includegraphics[width=0.75\columnwidth]{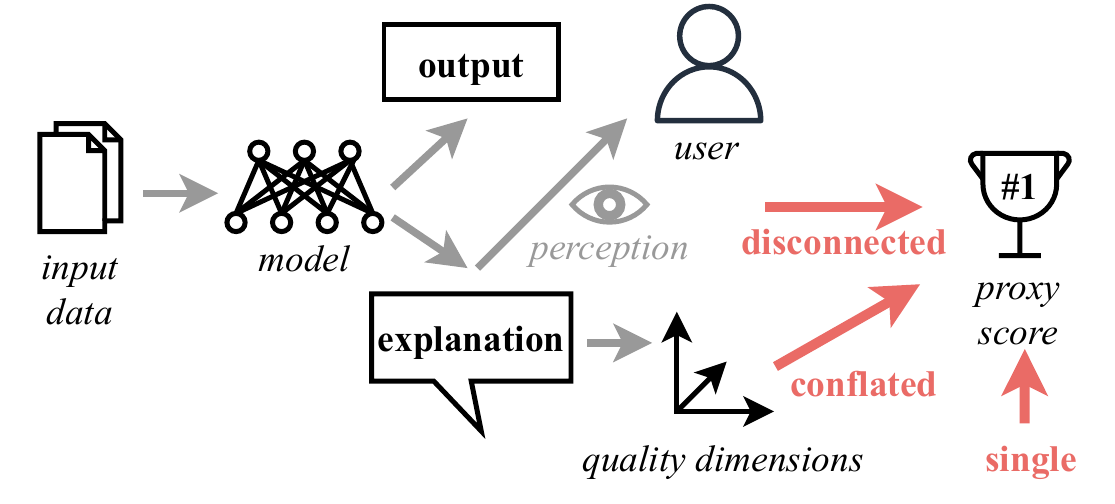}
\vspace{0.1cm}
\caption{Overview of the main drawbacks of current evaluation practices: (i) disconnect of proxy scores and user perception, (ii) conflation of multiple dimensions into single proxy scores, and  (iii) single-score leaderboards.}
\label{fig:problem_overview}
\end{figure}

While deep neural network models, such as transformers, achieve state-of-the-art results on many natural language processing (NLP) tasks, they are mainly black-box models. 
This raises the need to provide---in addition to the systems' predictions---explanations of \textit{why} the systems made these predictions.
Such explanations are essential when deploying models in real-world scenarios with human end-users \citep{angwin2016machine,Rudin2020Ageof} as they can help to identify dangerous model behavior and thereby, e.g., prevent discrimination.
Explanations can be given in diverse forms.
A popular type of explanation shows which parts of the model input were necessary for the model's output.
It is frequently implemented as a heat map over the input that visualizes an importance measure, such as integrated gradients \citep{DBLP:conf/icml/SundararajanTY17} or attention weights \citep{wiegreffe-pinter-2019-attention}.
Alternatively, model behavior can also be explained by providing supporting arguments for the model's prediction, e.g., by selecting a set of facts from a context \citep{yang-etal-2018-hotpotqa} or by generating textual explanations that directly verbalize why a model took a decision \citep{camburu_e-snli_2018}.
\Cref{tab:explanation_types} shows examples of these three types of explanations for the example of text-related tasks.

\begin{table}[t]
    \centering
    \resizebox{0.95\textwidth}{!}{%
    \begin{tabular}{C{0.14\columnwidth}L{0.19\columnwidth}L{0.35\columnwidth}L{0.33\columnwidth}}
        \toprule
         \textbf{Type} & \textbf{Description} & \textbf{Example} & \textbf{Proxy Scores} \\
         \midrule
         \textbf{Rationals, saliency maps} & Input tokens are highlighted to reflect what was most important to the model (e.g., based on attention or saliency scores such as integrated gradients). & \textit{Input}: I like this movie. \hspace{1cm} \textit{Prediction}: positive sentiment\vspace{0.1cm} \fbox{\parbox{0.33\columnwidth}{\raggedright\textit{Rational explanation}:\hspace{2cm}I \underline{like} this movie.\vspace{0.1cm}\hspace{3cm}\textit{Saliency explanation}:\hspace{2cm}I \colorbox{c_255_74_74}{like} this \colorbox{c_255_197_197}{movie}.}} &  \textbf{Overlap to human rational annotations} (e.g., via F1),\vspace{0.1cm} \hspace{2cm} \textbf{Removal analysis} (e.g., quantifying the drop in performance when highlighted input parts are removed) \citep[i.a.,][]{atanasova-etal-2020-diagnostic},\vspace{0.1cm} \hspace{1cm}  \textbf{Student model accuracy gains} when trained on the explanations \citep{pruthi-etal-2022-evaluating}\\
         \midrule
         \textbf{Supporting facts} & A set of facts (i.e., sentences) extracted from a given context are provided as evidence for the prediction. & \textit{Question}: What is the area of the desert that Ghanzi is in the middle of? \hspace{2.5cm} \vspace{0.1cm} \textit{Answer}: 900000 km² \vspace{0.1cm} \fbox{\parbox{0.33\columnwidth}{\raggedright Fact 1: Ghanzi is a town in the middle of the Kalahari Desert the western part of the Republic of Botswana in southern Africa. \hspace{3cm} Fact 2: The Kalahari Desert is [...] extending for 900000 km².}} & \textbf{Overlap to human annotations} of supporting facts (e.g., via F1) \citep{yang-etal-2018-hotpotqa},\vspace{0.1cm} \hspace{2cm} \textbf{Removal and consistency analysis} \citep{schuff-etal-2020-f1} \\
         \midrule
         \textbf{Free text} & Generated textual explanation that supports the prediction. & \textit{Premise}: A man in an orange vest leans over a pickup truck. \hspace{1cm} \textit{Hypothesis}: A man is touching a truck. \hspace{4cm}\textit{Predicted label}: entailment \textit{Explanation}: \fbox{\parbox{0.33\columnwidth}{Man leans over a pickup truck implies that he is touching it.}}\vspace{0.1cm} (from \cite{camburu_e-snli_2018}) & \textbf{Overlap to human-written references} (e.g., via BLEU or BLEURT) \citep{camburu_e-snli_2018,schuff-etal-2021-external,DBLP:conf/iccv/KayserCSEDAL21}) \\
         \bottomrule
    \end{tabular}%
    }
    \vspace{0.3cm}
    \caption{Three examples of different explanation types in NLP along with proxy scores that are used to quantify their quality.}
    \label{tab:explanation_types}
\end{table}

The predominant approach to evaluating the produced explanations' quality is to use automatic proxy scores, such as BLEU or F1 \citep{camburu_e-snli_2018,yang-etal-2018-hotpotqa,deyoung-etal-2020-eraser,atanasova-etal-2020-diagnostic}.
These scores are, in turn, ranked on leaderboards to define state-of-the-art systems.
Much research goes to great lengths to reach new highscores on these leaderboards.
However, it is still to be determined if developing systems that provide explanations that reach new highscores will also produce systems that provide a larger benefit to their human users.

This paper argues that the prevalent explanation evaluation process fails to measure explanation quality in a meaningful way.
We first \textbf{discuss desired characteristics of explanation quality} based on findings from social sciences, such as \citet{miller_explanation_2019}.
Next, we \textbf{show that current evaluation practices conflict with these characteristics}.
We support our argumentation with initial \textbf{empirical evidence from a crowdsourcing study} in which we investigate explainable question answering systems from the HotpotQA \citep{yang-etal-2018-hotpotqa} leaderboard.\footnote{\url{https://hotpotqa.github.io/}}
Concretely, we demonstrate the lack of proxy score validity, the corresponding conflation of quality dimensions and the erosion of target scores over time (i.e., \textit{Goodhart's Law}).

Finally, we propose practical guidelines to overcome those obstacles to meaningful explanation quality evaluation.
We release the collected human ratings and the corresponding analysis code.\footnote{\url{https://released-upon-acceptance.com} (we remove questions and answer texts to prevent leakage of the secret test set)}

\section{Characteristics of Explanation Quality}\label{sec:characteristics}
Criteria for high-quality explanations have mainly been discussed in social sciences so far.
Besides desirable explanation features such as coherence \citep{thagard_explanatory_1989,ranney_explanatory_1988, read_explanatory_1993}, soundness or completeness \citep{kulesza_too_2013}, literature has pointed out the importance of the explainees \citep{miller_explanation_2019,DBLP:conf/iui/WangY21}
and their goals \citep{vasilyeva_goals_2015}.
Based on this prior work, we discuss characteristics of explanation quality in NLP in this section.
Note that we assume the faithfulness of an explanation and only focus on characteristics for its \emph{perceivable quality}.\footnote{We consider explanation characteristics that can be judged without access to the underlying model.
We refer to \citet{jacovi-goldberg-2020-towards} for a discussion of faithfulness evaluation and to \cite{liao2022connecting} for a distinction between model-intrinsic and human-centered explanation properties.
}

\subsection{Explanation Quality is User-dependent}\label{sec:c_explainees}
We argue that in AI, an explanation exists only in relation to a system that should be explained (the \textit{explanandum}) and the human that receives the explanation (the \textit{explainee}).
This statement is in line with the social process function of an explanation described by \cite{miller_explanation_2019} referring to the conversational model of explanation of \cite{hilton1990conversational}.
Hilton argues that an explanation should be considered a conversation and emphasizes that \textbf{``the verb to explain is a three-place predicate: Someone explains something to someone''} \citep{miller_explanation_2019}.
Given that explanations are always targeted towards a specific user group, we argue that their quality needs to be assessed accordingly.

In the following, we detail how user goals, individual user characteristics as well as general peculiarities of human perception impact the definition of how \textit{good} an explanation is and why such a definition can never be universal.

\paragraph{Goals of Target Users}
\cite{vasilyeva_goals_2015} showed that users' perception of explanation quality depends on their goals.
Similarly, \cite{liao2022connecting} found that users' usage context affects which explanation quality properties they consider to be important.

While, for example, an explanation in the form of a heatmap over a text (as shown in the first row of \Cref{tab:explanation_types}) might be sufficient for an NLP developer or researcher who aims at analyzing and improving the system, it might not fit the needs of an end-user who has no machine-learning background but uses the system in practice.
Although the explanation contains the same information, its perceived quality might be considered lower by end-users compared to developers because, for example, the mental effort to process the explanation could be higher for end-users that are unfamiliar with such visualizations.

\paragraph{Individual Differences of Target Users} 
In addition to the users' goals, their background knowledge affects which type and extent of explanations are most useful for them \citep{,DBLP:journals/corr/abs-1810-00184,DBLP:journals/vi/YuS18,10.1145/3411764.3445088}.
As a trivial but illustrative example, a perfect explanation in Spanish is clearly useless to a monolingual English speaker and an ``explanation'' as it is provided by the coefficients of a linear model is useless to a user with dyscalculia.
Concretely, prior work showed that, i.a., (a) an increase in users' education levels and technical literacy corresponds to an increased algorithm understanding \citep{DBLP:conf/chi/ChengWZOGHZ19}, (b) users' need for cognition (i.e., their motivation to engage in effortful mental activities) impacts how much they benefit from interventions that increase analytical engagement with explanations \citep{DBLP:journals/pacmhci/BucincaMG21}, and (c) the effect of explanation on users strongly depends on the users' domain knowledge \citep{DBLP:conf/iui/WangY21}.

\paragraph{Intersubjective Quality within User Groups}
While the individual goals and characteristics of each user make them perceive and use explanations in a unique way, certain groups of ``similar'' explainees (e.g., Spanish native speakers reading a generated Spanish text) will be affected by explanations similarly.
Therefore, we argue that explanation quality is an \textit{intersubjective} construct.
This has two immediate implications.
First, it implies that every evaluation of explanation quality is limited to a specific group of explainees.
However, it also implies that explanation quality can be \textit{objectively} assessed within a suitable group of explainees.
For example, an often-used categorization in explainability is to divide users into three groups: developers, domain experts and lay users \citep{DBLP:conf/iui/RiberaL19}.
Dividing users into such high-level groups can already help to identify important differences regarding their explanation needs, however a more fine-grained categorization including, e.g., social and cognitive user properties---as suggested by \cite{DBLP:journals/corr/abs-2201-11239}---could further improve evaluation quality.

\paragraph{Cognitive Biases and Social Attribution}
Hilton's conversational conversational model of explanation distinguishes two stages: (a) the diagnosis stage in which causal factors of an event/observation are determined and (b) the explanation presentation stage in which this information is communicated to the explainee \cite{hilton1990conversational}.\footnote{We refer to \cite{miller_explanation_2019} for a more detailed discussion of Hilton's work in the context of explainability.}
So even if the first stage is successful (i.e., the right ``explanation information'' has been identified), \textit{communicating} the explanation information can fail (e.g., by relying on an inappropriate visualization to visualize the information).
\citet{10.1145/3531146.3533127} empirically showed that such problems in explanation communication can occur for heat map explanations over text and that the information that users understand from these explanations is distorted by unrelated factors such as word length.
Similar, \citet{DBLP:conf/acl/GonzalezRS21} show that belief bias (i.e., a particular cognitive bias) affects which explanation method users prefer.
More broadly, \cite{DBLP:journals/corr/abs-2201-11239} propose a framework of social attribution by the human explainee that describes which information an explainee is comprehending from an explanation and thereby allows to identify failures of explainability methods.

\subsection{Explanation Quality has (Orthogonal) Dimensions}\label{sec:c_multidimensional}
Explanation quality is commonly treated as a monolithic construct in which explanations can be ranked along a unidimensional range of explanation ``goodness''.
We, in contrast, argue that there are different dimensions of explanation quality and that these dimensions can be orthogonal to each other.
Thus, explanation should be evaluated along \textit{multiple} facets of explanation quality.

An example of two orthogonal quality dimensions are faithfulness and plausibility.
Consider an explanation that explains the decision process of a system A in a way that (a) faithfully reflects the system decision process and (b) plausibly convinces a user of the correctness of the prediction.
We then replace the system with a new system B while keeping the explanation constant.
The explanation will still be plausible to the user (it did not change), however, if system B has a different decision process, the explanation is not faithful anymore as it no longer reflects the model's inner workings.
Consequently, the two explanation quality dimensions faithfulness and plausibility can be independent and cannot be captured with the same score.\footnote{We refer to \cite{jacovi-goldberg-2020-towards} for a detailed comparison of faithfulness and plausibility.}

Similarly, an explanation that is\textit{ perceived to be helpful} by explainees does not actually have to \textit{be helpful} for them.
\cite{DBLP:conf/iui/BucincaLGG20} showed that between two decision support systems, users preferred one system (in terms of rating it as more helpful and trusted), although their actual performance was significantly better with the less-favored system.
Also, their subjective ratings were not predictive of their objective performance with the system.
In their follow-up work, \cite{DBLP:journals/pacmhci/BucincaMG21} found a trade-off between subjective system quality ratings and effective human-AI performance for explainable AI systems.
Related effects have been reported by \cite{scharrer2012seduction} who compared the impact of showing easy versus difficult scientific arguments to lay people and found that the easy arguments lead to participants being more convinced and underestimating their own knowledge limitations.
These findings suggest that effective explanations have to combine or balance (a) their \textit{perceived} utility and (b) their \textit{actual} utility to their users.
While an explanation that only subjectively seems to provide a benefit clearly is not desirable, an explanation that affects users to their own benefit but is disliked by them will not be used in practice.\footnote{\cite{nadarzynski2019acceptability} found AI acceptability to be correlated with, i.a., perceived utility and trustworthiness.}

Overall, effective explanation evaluation thus has to account for the numerous, partially orthogonal dimensions of explanation quality.

\section{Case Study on the HotpotQA Leaderboard}\label{sec:case_study}
\begin{figure}
    \centering
    \includegraphics[width=\textwidth]{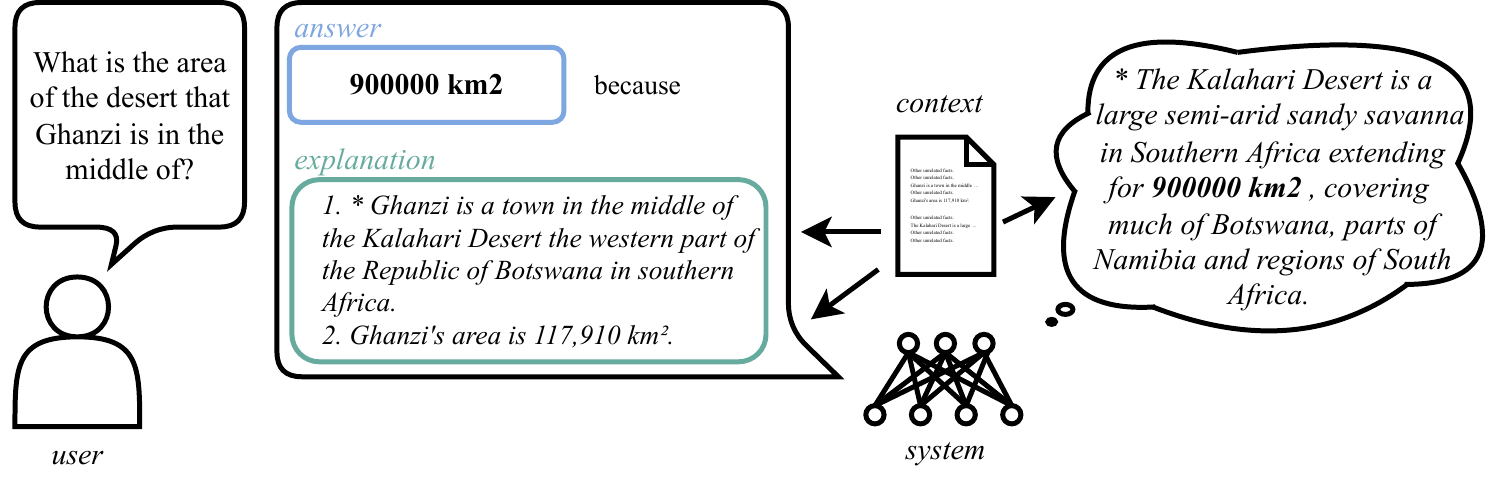}
    \caption{Example prediction from the HotpotQA explainable question answering dataset. The model resturns the correct answer (blue box) but its predicted explanation, i.e., selection of supporting facts (green box), is only partially correct as it (a) reports an irrelevant fact about the size of Ghanzi and (b) fails to report the relevant fact containing the predicted answer. ``$\ast$'' marks facst within the human-annotated ground truth explanation. How can the resulting (lack of) explanation quantity be evaluated meaningfully? Figure adapted from \cite{schuff-etal-2020-f1} showing example data from \cite{yang-etal-2018-hotpotqa}.}
    \label{fig:hotpot_example}
\end{figure}

Before we present our arguments on how current explainability quality evaluations fall short, we introduce our case study which we will refer back to throughout \Cref{sec:problems}.
To support the following discussion with empirical evidence, we conduct a crowdsourcing study analyzing systems from \textbf{10 real models} submitted to the official HotpotQA \citep{yang-etal-2018-hotpotqa} leaderboard ranking explainable question answering models.\footnote{We thank the HotpotQA maintainers for providing us with the predictions and the system submitters for giving us their consent to include their model in our case study.}

\subsection{Task, Models and Automatic Evaluation}\label{sec:task_model_evaluation}
In the following, we describe the HotpotQA task, present the leaderboard models we analyze and list proxy metrics that we use to automatically quantify the models' explanation capabilities.

\paragraph{Task and Data}
In the HotpotQA task, systems receive a question and parts of 10 Wikipedia\footnote{In our case study, we focus on the distractor setting.} articles as context and have to predict (i) an answer to the question (yes/no or a text span from the context) as well as (ii) which sentences from the context are \textit{supporting facts} to their predicted answer.
The supporting facts serve as an explanation for the predicted answer of the model.
The HotpotQA dataset provides gold annotations for answers as well as supporting facts for 113k instances in total.
\Cref{fig:hotpot_example} shows an example of a (real) model prediction that contains the correct answer and one of two ground facts but misses the second ground truth fact and, instead, returns a fact that is unrelated to the answer.
The training and development splits of the dataset are publicly available while the test set is only used to evaluate models on the official leaderboard.\footnote{\url{https://hotpotqa.github.io/}}

\paragraph{Evaluated Models}
We obtained consent from submitters of 24 \textbf{real models} to include the system predictions in our analysis.
From those 24 models, we choose 10 models for our user study: AMGN (rank 16) (anonymous submittor), FE2H on ALBERT (3) \citep{DBLP:journals/corr/abs-2205-11729}, HGN \citep{DBLP:conf/emnlp/FangSGPWL20} (35), IRC (63) \citep{DBLP:conf/ijcnn/NishidaNSY21}, Longformer (25) (anonymous), S2G-large (31) (anonymous), Text-CAN (47) (Usyd NLP), GRN (65) (anonymous), SAE (48) \citep{DBLP:conf/aaai/TuHW0HZ20}, DecompRC (unranked\footnote{DecompRC reports answer metrics only.}) \citep{DBLP:conf/acl/MinZZH19}.\footnote{Ranks last updated 24th of February 2023.}

Additionally, we derive\textbf{ five synthetic models} using the ground truth annotations to include extreme cases of the potential space of systems:
(i) \textit{gold answers and gold facts} (plain gold annotations),
(ii) \textit{gold answers and random facts} (we sample the same number of of facts as the gold annotations, but do not sample from the articles in which the gold facts are located in),
(iii) \textit{random answers and gold facts} (we sample a random answer from the context while keeping the number of words the same as in the gold answer),
(iv) \textit{random answers and random facts} (both answers and facts are sampled, as described before),
(v) \textit{gold answers and all facts} (gold answers but the predicted facts are \textit{all} facts from the context, i.e. from 10 Wikipedia articles).

\paragraph{Proxy Scores}

\begin{table}[t]
    \centering
    \resizebox{0.95\textwidth}{!}{%
    \begin{tabular}{C{0.2cm}C{0.3\columnwidth}p{0.5\columnwidth}}
        \toprule
          & \textbf{Proxy score} & \textbf{Description}  \\
         \midrule
         \multirow{10}{*}{\rotatebox[origin=c]{90}{\textit{leaderboard scores}}} & \textbf{answer-precision, answer-recall, answer-F1, answer-EM} & Overlap metrics that compare the predicted answer tokens and the ground truth answer tokens using precision, recall, F1 and exact match (EM) \\
         & \textbf{SP-precision, SP-recall, SP-F1, SP-EM} & Overlap metrics that compare the set of predicted supporting facts and the set of ground truth supporting facts using precision, recall, F1 and EM on a sentence level   \\
         & \textbf{joint-precision, joint-recall,} \fbox{\textbf{joint-F1}}\textbf{, joint-EM} & Joint versions of the answer and supporting facts metrics based on instance-wise products of EM, precision and recall \\
         \midrule
         \multirow{8}{*}{\rotatebox[origin=c]{90}{\textit{additional scores}}} & \textbf{LocA score} &  A score that measures how well the predicted answer and explanation are coupled. It compares the fraction of answer tokens inside an explanation to the fraction of tokens outside an explanation.\\
         & \textbf{\#facts} & Number of facts (i.e., sentences) within the predicted explanation \\
         & \textbf{\#words} & The number of words over all facts inside the predicted explanation \\
         \bottomrule
    \end{tabular}%
    }
    \vspace{0.3cm}
    \caption{Proxy scores that we use to automatically evaluate the explainable question answering systems. The upper part shows the scores that the HotpotQA leaderboard evaluates. The lower part shows additional metrics that are (a) two simple surface metrics related to the length of the predicted explanation and (b) one task-specific explanation quality score. \fbox{joint-F1} is used to rank models on the leaderboard.}
    \label{tab:proxy_scores}
\end{table}

The HotpotQA leaderboard reports the metrics exact match (EM), precision, recall and F1 for three levels: (i) answer, (ii) supporting facts (i.e., the explanation) and (iii) on the answer and explanation jointly.
\Cref{tab:proxy_scores} lists and describes all of these proxy scores in the upper part of the table.
The leaderboard ranks the systems according to joint-F1 scores on a non-public test set (breaking ties by using other measures like joint-EM and answer-F1).

We consider three additional scores shown in the lower part of \Cref{tab:proxy_scores}.
The LocA score is a task-specific score proposed by \cite{schuff-etal-2020-f1} and measures to which extent predictions and explanations are coupled.
LocA is defined as
\begin{equation*}
    \textsc{LocA} := \frac{\frac{I}{A}}{1 + \frac{O}{A}} =  \frac{I}{A + O} \in [0,1]
\end{equation*}
where $I$ and $O$ denote the number of answers inside/outside of the predicted facts and $A$ denotes the total number of answers.
A higher LocA score corresponds to a better explanation-answer coupling.
While the original LocA score measures how many predicted answers are located in explanations by comparing offsets, we generalize this concept to general string matching because we do not have access to the answer offsets of the leaderboard models.
As a positive side-effect, this makes the score applicable to every kind of model and not only to extractive question answering models.

Furthermore, we include two additional surface scores that measure an explanation's length in terms of (a) the number of facts it includes and (b) the total number of words that these facts contain.

We provide all F1, LocA and explanation length values of the 24 models for which we got permission to include them in our analysis as well as our five synthetic models in \Cref{tab:leaderboard} in \Cref{sec:detailed_proxy_scores}.

\subsection{Human Evaluation}\label{sec:case_study_human_evaluation}
To obtain a clearer perspective onto (i) the relation between the described proxy scores and human ratings and (ii) the model ranks regarding various human ratings, we analyze \textit{test set predictions} of the 10 \textit{real} model submissions as well as the five synthetic models we discussed in \Cref{sec:task_model_evaluation}.
We evaluate the models in a crowdsourced user study with 75 participants, collecting subjective quality ratings of \textbf{utility}, \textbf{consistency}, \textbf{usability}, \textbf{answer correctness} and \textbf{mental effort} as well as objective \textbf{completion time} measures.

Note that, although \citet{schuff-etal-2020-f1} already conduct a human evaluation to investigate the relation between the different proxy scores and various human ratings for HotpotQA, their evaluation is limited to three models and the ground truth predictions on the public validation set only.
In the following, we describe the conducted user study in more detail.

\paragraph{Experiment Design and Participants}
We make use of a between-subject experiment design, i.e., each participant is exposed to model predictions from exactly one model.
The participants are distributed to models such that each model receives ratings from five different participants.\footnote{We ensure that each participant only participates once across the whole experiment.}
We include two attention checks to filter out participants that do not read the question or the explanations.

For each model, we collect ratings from five crowdworkers who each rate a sample of 25 questions drawn from a pool of 100 questions.\footnote{To support our assumption that a pool of 100 questions is sufficiently representative, we simulate experiments with various question subsets.
We find that correlations stabilize for as few as 20 questions and report details in \Cref{sec:few_questions_simulation}.}
For each participant, we present the individual sample of 25 questions in a randomized order to avoid potential carry-over effects between questions.
We make use of this approach to (i) cover a large amount of questions to better reflect the dataset and at the same time (ii) restrict the user's workload to evade fatigue effects.

We recruit a total of 75 crowdworkers from the US using Mechanical Turk.
We require workers to have a $>$90\% approval rate and an MTurk Master qualification and ensure that each worker participates no more than once in our experiments.

\paragraph{Collected Human Ratings}
\begin{table}[t]
    \centering
    \resizebox{0.95\textwidth}{!}{%
    \begin{tabular}{p{0.10cm}C{0.25\columnwidth}p{0.6\columnwidth}}
        \toprule
          & \textbf{Quality dimension} & \textbf{Description}  \\
         \midrule
         \multirow{9}{*}{\rotatebox[origin=c]{90}{\textit{instance}}} & \textbf{Explanation \mbox{utility}} & ``The explanation helps me to decide if the answer is correct.'' \\
          & \textbf{Explanation \mbox{consistency}} & ``The explanation helps me to understand how the model came up with its answer'' (similar to \citet{Nourani-Kabir-Mohseni-Ragan-2019} and \citet{schuff-etal-2020-f1}) \\
          & \textbf{Answer \mbox{correctness}} & ``The answer is correct'' (similar to \citet{DBLP:conf/ichi/BussoneSO15}, \citet{camburu_e-snli_2018},  \citet{schuff-etal-2020-f1}, \citet{kumar-talukdar-2020-nile}, and \citet{DBLP:journals/corr/abs-2004-14546})\\
          & \textbf{Completion time} & Time per instance (similar to \cite{DBLP:conf/chi/LimDA09}, \cite{Lage-Chen-He-Narayanan-Kim-Gershman-Doshi_Velez-2019}, \cite{DBLP:conf/chi/ChengWZOGHZ19}, and \cite{schuff-etal-2020-f1}) \\
         \midrule
         \multirow{2}{*}{\rotatebox[origin=c]{90}{\textit{system}}} & \textbf{Usability} & UMUX system usability questionnaire \citep{finstad_usability_2010,finstad_response_2013}. \\
         & \textbf{Mental effort} & Paas mental effort scale \citep{paas1992training}\vspace{0.08cm}\\
         \bottomrule
    \end{tabular}%
    }
    \vspace{0.3cm}
    \caption{Human ratings/scores collected in our crowdsourcing study.}
    \label{tab:human_scores}
\end{table}

We collect the human ratings/scores listed in \Cref{tab:human_scores}.
We collect \textit{per-instance} participant ratings of perceived \textbf{explanation utility}, \textbf{explanation consistency} and \textbf{answer correctness}.
In addition, we track the \textbf{completion time}, the participants take to finish each question.
Further, we collect \textit{per-system} ratings within a post questionnaire at the end of the experiment where we ask participants to rate \textbf{usability} using the UMUX scale \citep{finstad_usability_2010,finstad_response_2013} and \textbf{mental effort} using the Paas scale \citep{paas1992training}.

\paragraph{Results}
We discuss our results and, in particular, the relation between the proxy scores and the collected human ratings in the context of the respective shortcomings in the evaluation of explanation quality in the following section.
We provide the detailed averaged ratings over all 15 models in \Cref{automatic_scores_and_human_ratings} (\Cref{tab:human_ratings}).
In order to ease readability, \fcolorbox{gray}{gray!8}{we highlight the main results of our case study using grey boxes}.
Further details on the collected proxy scores over all 24 real and 5 synthetic models are provided in \Cref{tab:leaderboard} in \Cref{sec:detailed_proxy_scores}.
Details on the exact human ratings over all models included in our human evaluation are provided in \Cref{tab:human_ratings} in \Cref{sec:human_rating_details}.

\section{Shortcomings of Current Evaluations}\label{sec:problems}
Explanation evaluation in NLP is mainly performed automatically \citep{yang-etal-2018-hotpotqa,deyoung-etal-2020-eraser,atanasova-etal-2020-diagnostic}, borrowing proxy scores from other tasks, such as accuracy, F1, BLEU \citep{papineni-etal-2002-bleu} or BLEURT \citep{sellam-etal-2020-bleurt}.
\citet{deyoung-etal-2020-eraser}, for example, propose to use automatic measures to evaluate explanation quality aspects, such as faithfulness, comprehensiveness and sufficiency.
While they compute a variety of automatic scores that capture (a) how faithful a model's predictions are (which they operationalize using comprehensiveness and sufficiency measures) and (b) how similar a model's predictions are to human-annotated rationales, their evaluation does not include a human evaluation of the model explanations and leaves open how their proposed measures relate to user-perceived explanation characteristics.
Other works evaluate explanation quality more indirectly, such as \citet{pruthi-etal-2022-evaluating}, who evaluate saliency explanations via the accuracy gains they provide when training a student model on the produced explanations.

In the following, we present common evaluation practices and assess to which extent they conflict with the explanation quality characteristics presented in \Cref{sec:characteristics}.
Figure \ref{fig:problem_overview} provides an overview of the main challenges discussed in this paper.

\subsection{Unvalidated Proxy Scores}\label{sec:p_unexplored_proxy_scores}
The underlying assumption of using proxy scores for evaluating explanation quality is that an improvement in proxy scores implies an increase in user benefits.
However, to the best of our knowledge, there is no established view to which extent those scores actually reflect the value of explanations to users (i.e., to which extent they are \emph{valid} and measure what they should measure).
This practice conflicts with both, the user-centered (\Cref{sec:c_explainees}) and the multidimensionality characteristic (\Cref{sec:c_multidimensional}) of explanation quality.
In the following, we discuss different aspects of the relation between proxy scores and human ratings.
Concretely, we investigate (i) the pairwise relation between proxy scores and human ratings, (ii) their overall relation in terms of their underlying factor structure and (iii) the dynamics of their relation over time.

\subsubsection{Low Correlations Between Proxy Scores and Human Ratings}
If the assumption that higher values on a specific proxy score correspond to higher user benefits holds true, this benefit should also reflect in one or multiple human subjective ratings or objective performance markers.

One of the few studies that study the strength of the link between proxy scores and human ratings is conducted by \cite{DBLP:conf/iccv/KayserCSEDAL21}. 
They analyze the correlation between natural language generation (NLG) metrics (i.a., BLEU, BERT-Score and BLEURT) and human quality ratings to quantify free text explanations for three visual question answering datasets within a large crowdsourcing study.
They find that (averaged over the three datasets), the highest Spearman correlation across 10 different proxy scores only reaches 0.29.
Similarly, earlier work of \cite{camburu_e-snli_2018} finds that BLEU does not reliably reflect human quality ratings of textual explanations for a natural language inference (NLI) task which is supported by the results of \cite{schuff-etal-2021-external} who find that even high performance differences regarding proxy scores of explainable NLI models are not reflected in human ratings of explanation quality.
\cite{clinciu-etal-2021-study} study human ratings of explanation clarity and informativeness in the context of natural language explanations of Bayesian networks.
Averaged over different scenarios, they find that, across 11 proxy scores, the highest Spearman correlation still only reaches 0.39 while the correlation between the two human ratings reaches a much higher value of 0.82.

Overall, to the best of our knowldege, all (of the few) available studies indicate that \textbf{proxy scores and human ratings correlate weakly}.

\begin{figure*}[t]
    \centering
    \begin{subfigure}[t]{.38\textwidth}
        \centering
    \includegraphics[width=\textwidth]{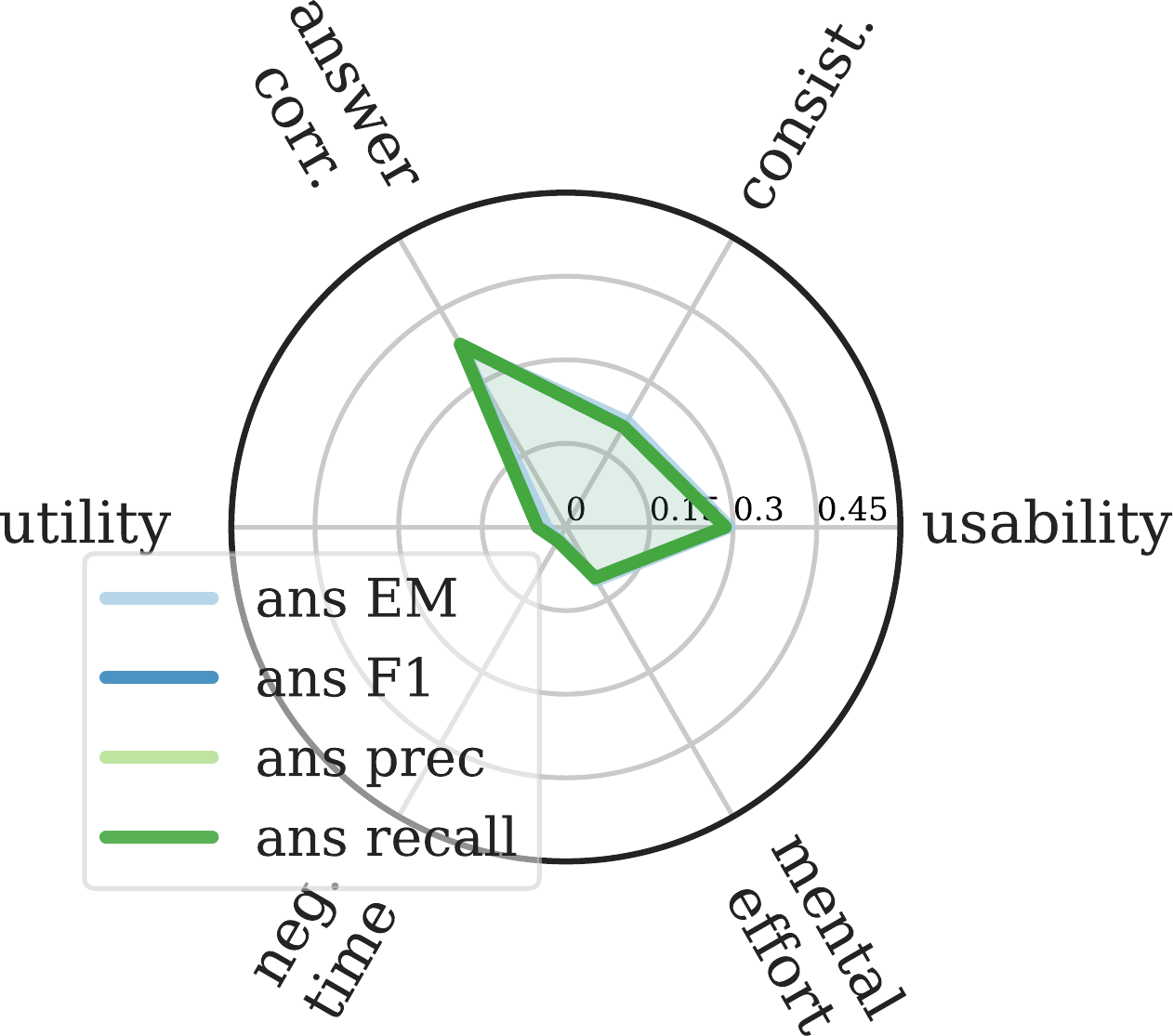}
    \end{subfigure}%
    \hspace{0.1\textwidth}
    \begin{subfigure}[t]{.38\textwidth}
        \centering
    \includegraphics[width=\textwidth]{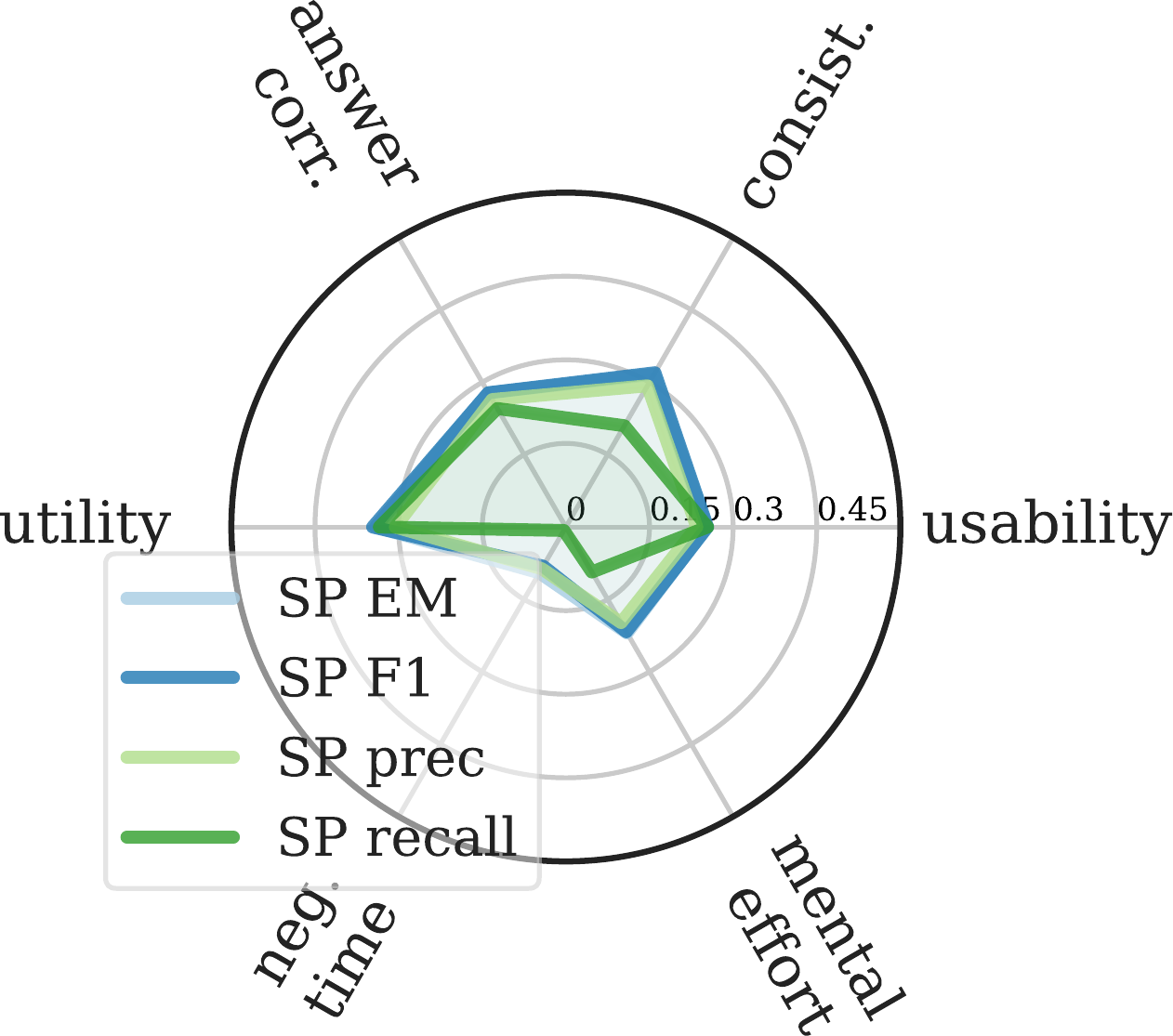}
    \end{subfigure}
    %\hspace{0.01\textwidth}
    \begin{subfigure}[t]{.38\textwidth}
    \vspace{0.1cm}
        \centering
    \includegraphics[width=\textwidth]{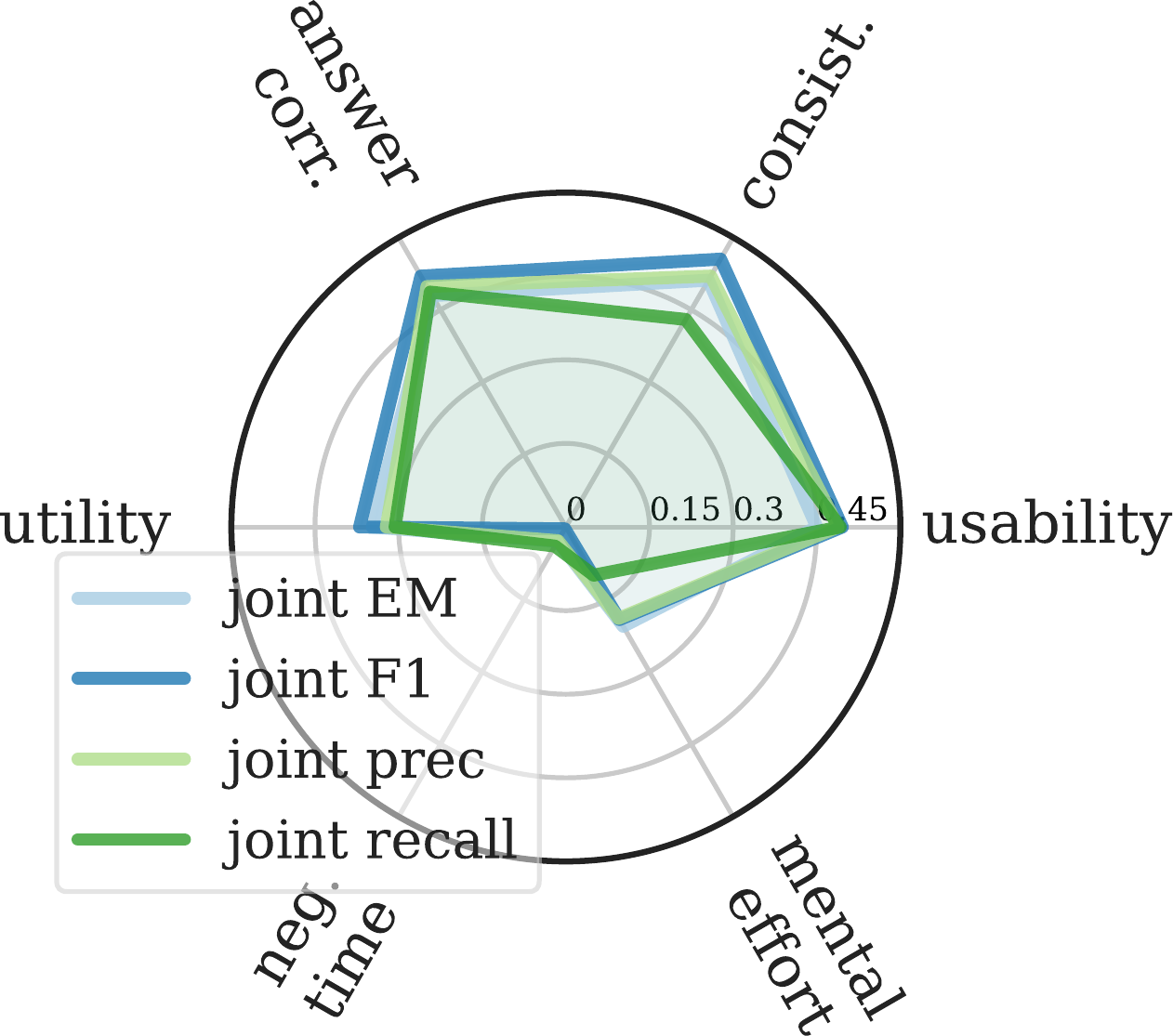}
    \vspace{0.1cm}
    \end{subfigure}%
    \hspace{0.1\textwidth}
    \begin{subfigure}[t]{.38\textwidth}
    \vspace{0.1cm}
        \centering
    \includegraphics[width=\textwidth]{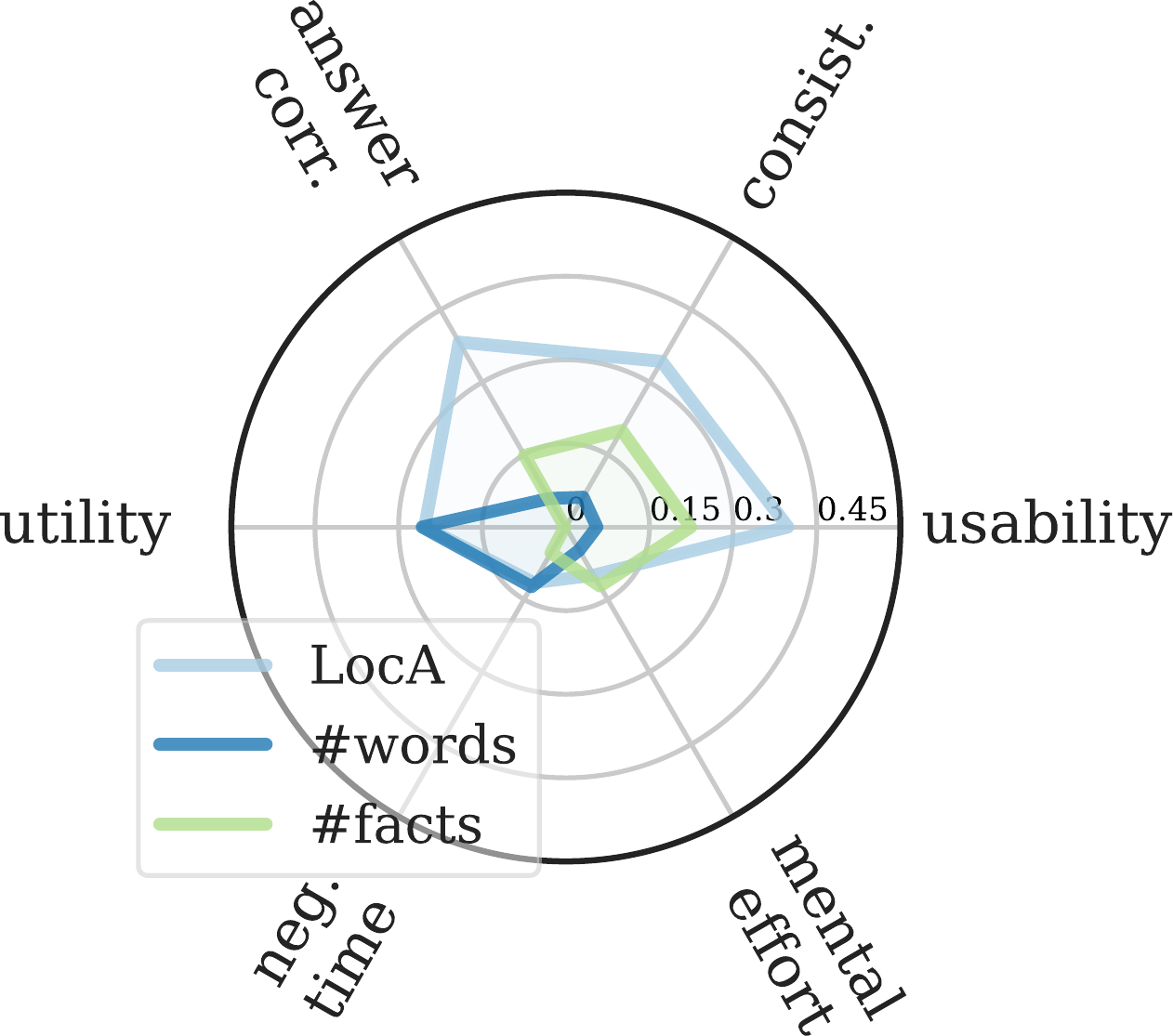}
    \vspace{0.1cm}
    \end{subfigure}%
    \caption{Kendall's $\tau$ correlation coefficients for the correlation of different automatic scores and user-rated quality dimensions. The correlations illustrate the weak and conflated connection between proxy scores and human assessment (from left to right and top to bottom: scores evaluating answer correctness, scores evaluating correctness of supporting facts, scores jointly evaluating answer and fact correctness, additional scores including LocA and surface scores). Axes cropped at 0.6.}\label{fig:radar_plots}
\end{figure*}

\paragraph{Case Study}
We provide an additional analysis for explainable question answering and exceed previous studies in the diversity of human rating dimensions.
Additionally, our results are evaluated using predictions of real leaderboard models instead of evaluating correlations on ground truth datasets \citep{camburu_e-snli_2018,DBLP:conf/iccv/KayserCSEDAL21,clinciu-etal-2021-study} and thus cover a distribution of scores and ratings that better reflects the scores' usage within leaderboards.

\Cref{fig:radar_plots} shows Kendall's $\tau$ correlation coefficients between (a) the automatic scores included in the leaderboard and (b) the human ratings we collected in our study.
A heat map visualization of all pairwise correlations including statistical significance markers and coefficients can be found in \Cref{fig:human_machine_correlations_kendall} in \Cref{automatic_scores_and_human_ratings}.

\vspace{0.2cm}
\casestudybox{
While we observe  moderate correlations between, e.g., joint-F1 and explanation consistency, the majority of correlations is under 0.5 and thus the previously described weak relation between human ratings and proxy scores is \textbf{supported by our case study}.
}

\subsubsection{Proxy Scores Conflate Different Dimensions}
We argue that the currently used explanation quality proxy scores can, and often will, conflate different dimensions of explanation quality, and, consequently, information about the individual independent dimensions is lost and cannot be recovered.
For example, given two systems with similar proxy scores, it cannot be determined which one was superior in terms of individual explanation quality aspects, such consistency or understandability.
Therefore, it is not possible to identify an isolated improvement of a model in some of those aspects using the proxy score.
For example, when we improve the proxy score, we cannot assess whether we actually improved all quality aspects or only a subset of them (and possibly decreased the performance on others).
Similarly, a targeted improvement of particular quality aspects (e.g., for a particular use-case) is not possible.

\begin{table}[t]
    \centering
    \begin{tabular}{L{0.2\columnwidth} C{0.07\columnwidth} C{0.05\columnwidth} C{0.05\columnwidth} C{0.07\columnwidth} C{0.05\columnwidth}}
        \toprule
          Score/rating & Type & F1 & F2 & F3 & F4 \\
         \midrule
            answer-EM        & \computeremoji  & 0.98 &      &      &      \\
            answer-F1        & \computeremoji  & 0.97 &      &      &      \\
            answer-recall    & \computeremoji  & 0.97 &      &      &      \\
            answer-precision      & \computeremoji  & 0.97 &      &      &      \\ 
            correctness rating & \judgeemoji & 0.80 &      &      &      \\ 
        \midrule
            sp-EM            & \computeremoji  &      & 0.97 &      &      \\
            sp-F1            & \computeremoji  &      & 0.89 &      &      \\
            sp-precision          & \computeremoji  &      & 0.83 &      &      \\ 
            sp-recall        & \computeremoji  &      & 0.82 &      &      \\
            joint-EM         & \computeremoji  &      & 0.58 &      &      \\
            joint-F1         & \computeremoji  &      & 0.55 &      &      \\
        \midrule
            consistency rating & \judgeemoji &      &      & 0.79 &      \\
            usability rating   & \judgeemoji &      &      & 0.74 &      \\
            joint-recall       & \computeremoji &      &      & 0.64 &      \\
            utility rating     & \judgeemoji &      &      & 0.63 &      \\
            LocA score         & \computeremoji &      &      & 0.56 &      \\ 
            joint-precision         & \computeremoji &      &      & 0.53 &      \\
            completion time    & \judgeemoji &      &      & (0.24) &      \\
        \midrule
            \#words          & \computeremoji   &      &      &      & 0.95 \\
            \#facts          & \computeremoji   &      &      &      & 0.95 \\ 
         \bottomrule
    \end{tabular}
    \vspace{0.3cm}
    \caption{Factor loadings of the collected scores/ratings onto four factors (F1-F4). Proxy scores are marked as \computeremoji, human ratings/scores are marked as \judgeemoji. We observe that F3 contains all explanation-related human ratings as well as the three three proxy scores joint-recall, LocA and joint-precision. This suggests that these three scores are better-suited to capture perceived explanation quality compared to the currently used joint-F1 which loads onto factor F2.}
    \label{tab:fa}
\end{table}

\paragraph{Case Study}
In order to assess the degree to which proxy scores are related to \textit{multiple} human ratings, we also analyze the ratings' factor structure.
While the correlations shown in \Cref{fig:radar_plots} demonstrates that the \textit{pairwise} correlations are weak, they do not tell us how a score/rating is associated to the ``big picture'' in terms of the latent structures behind the scores and ratings.
A factor analysis is used to describe each score/rating in terms of its association to empirically-derived latent factors.
In order to determine the number of factors, we follow the \textit{method agreement procedure} in which the optimal number of factors is chosen based on the largest concensus between numerous methods.
We find four dimensions to be supported by the highest number of methods as---among a total of 14 methods---four methods agree on a number of four factors (i.e., beta, Optimal coordinates, Parallel analysis and Kaiser) which has a higher support than every other number of factors (ranging between 1 and 19).
We therefore conduct a factor analysis using a varimax rotation that maps each score/rating to one of the four latent factors such that the resulting factors describe the data as good as possible.
\Cref{tab:fa} displays the respective loadings (i.e., correlations of the score/rating with the latent factor).

Factor F1 contains all answer-related scores including human ratings of answer correctness.
Factor F2 only contains the fact-related automatic scores as well as joint-F1 and joint-EM.
Factor F3 contains all explanation-related human ratings.
Interestingly, F3 also contains the automatic scores joint-recall, LocA and joint-precision.
Factor F4 contains the explanation-length-related automatic scores \#words and \#facts.

We observe that the answer-related proxy scores and the human correctness ratings form a cluster and have strong loadings on their joint factor.
This can be interpreted as evidence that perceived answer correctness can---to a moderately strong extent---be measured via the answer-related proxy scores.

If (one of) the evaluated proxy scores for explanation quality would have an equally strong association to any human rating, we would expect to observe a factor, that, e.g., contains joint-F1 and perceived utility along with strong factor loadings of both scores/ratings onto their joint factor.

However, \Cref{tab:fa} demonstrates that all of the different explanation-related human ratings can be found in one factor along with the proxy scores joint-recall, LocA and joint-precision.
This shows that our explanation quality measurements cannot be grouped into distinguishable groups (as we observe for the answer-related scores and ratings).
Instead, they form a hardly-interpretable diffuse factor 3 that mixes up all kinds of human ratings and yields much lower factor loadings and---in addition---does not contain the leaderboard ranking score joint-F1.

\vspace{0.2cm}
\casestudybox{
Overall, our factor analysis suggests that (a) the \textbf{answer-related proxy scores reflect human answer correctness ratings}, (b) \textbf{no explanation-related proxy score can be associated to a particular human rating}.
In particular, joint-F1 does not share a joint factor with any human rating.
}

\subsubsection{Goodhart's Law: Validity Can Change Over Time}\label{sec:goodhart}
Even if we had a score that is valid, i.e., it measures one dimension of explanation quality in a decent way, using this score as the sole ranking criterion of a leaderboard can subvert its validity over time.
This effect is described in \emph{Goodhart's Law} that is commonly stated as \textbf{``when a measure becomes a target, it ceases to be a good measure''} \citep{goodhart_problems_1975,campbell_assessing_1979,strathern_improving_1997, manheim_building_2018,DBLP:journals/corr/abs-1803-04585}. %\footnote{We refer to \citet{manheim_building_2018} for a clear and nuanced delineation of the underlying problems.}
\citet{thomas_reliance_2022} discuss this in the context of AI and highlight the field's problematic reliance on (single) metrics including the issue of metrics being gamed \citep{bevan_whats_2006}.

Assume that an initial investigation of some systems showed that a particular proxy score can be considered to be valid (in a certain use case for a certain user group).
If now more and more systems are developed with the primary goal of reaching higher values on that score, the initial set of models no longer represent the new model population.
As a result, it cannot be ensured that the original strong relation between the (initially valid) proxy score and the measured quality dimension still holds.
Consequently, the score's validity can ``wear off'' over time as it is used in isolation.

\begin{figure}[t]
    \centering
    \includegraphics[width=.8\textwidth]{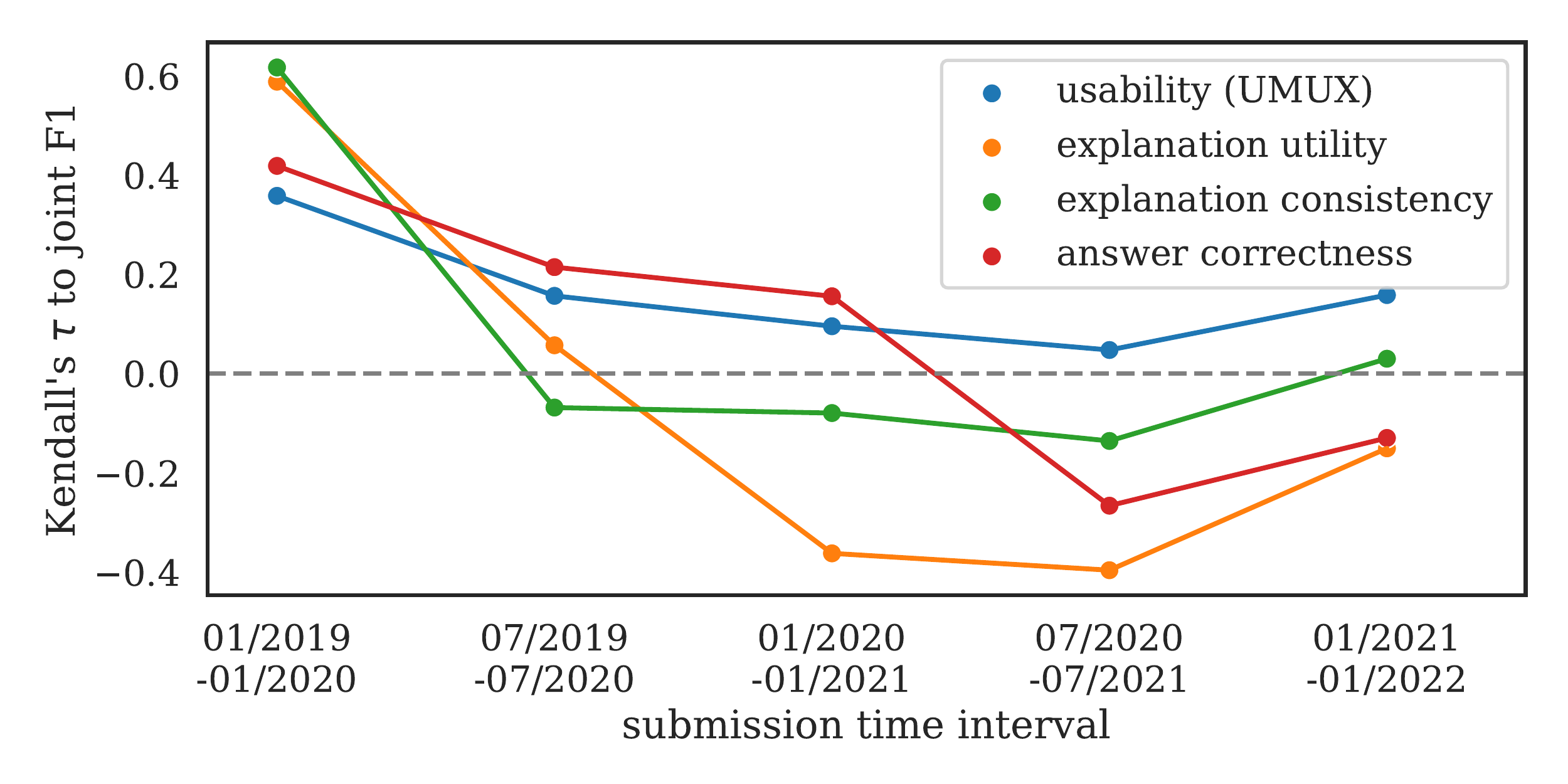}
    \vspace{0.1cm}
    \caption{Kendall's $\tau$ correlations over time between different human ratings and the official leaderboard metric joint-F1. The gradual decline of the relation between joint-F1 and human ratings indicates that joint-F1 looses validity over time and thus supports Goodhart's law.}
    \label{fig:correlations_over_time}
\end{figure}

\paragraph{Case Study}
We investigate whether we can find such a temporal deterioration in the HotpotQA leaderboard.
For this, we study the association of the leaderboard's target metric (i.e., joint-F1) with the measured human ratings across different time windows.
\Cref{fig:correlations_over_time} shows Kendall's $\tau$ correlation coefficients between joint-F1 and human ratings for a 12-month sliding window over system submissions.
We observe that correlations decrease from moderate positive to lower and even negative correlations.
We hypothesize that this decrease could have been mitigated using multiple proxy scores.
\vspace{0.2cm}
\casestudybox{
Overall, our observations indicate that Goodhart's law affects today's leaderboards and \textbf{single target metrics loose their expressiveness over time}.
}

\subsection{Neglecting Users}\label{sec:p_missing_user_studies}
\begin{figure}[t]
    \centering
    \includegraphics[width=0.8\textwidth]{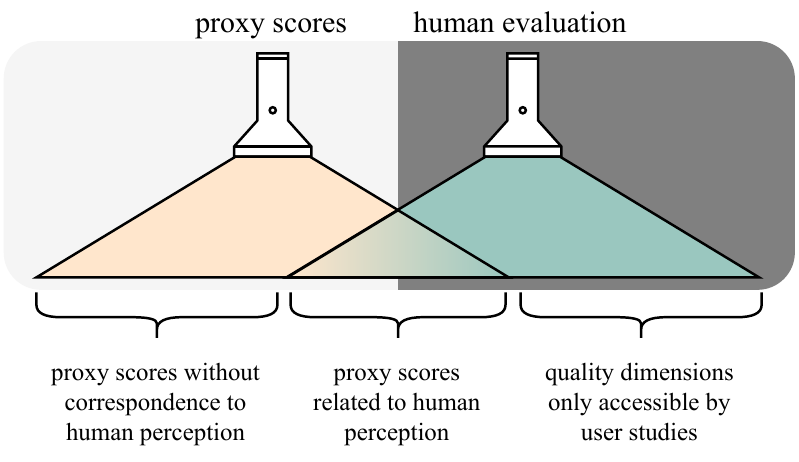}
    \caption{Visualization of the Streetlight Effect in the context of explanation quality evaluation. Searching where the light is, i.e., relying on proxy scores in isolation (left) does not allow to study the whole spectrum of explanation quality. Instead, we also have to study quality dimensions that are only accessible via human evaluation (right).}
    \label{fig:streetlight_effect}
\end{figure}

So far, we argued why the currently used proxy scores of explanation quality do not reliably reflect user-perceived quality properties.
But if we had proxy scores that were shown to successfully reflect various aspects of explanation quality, could we stop human evaluation and rely on these scores alone?
We, just like \citep{thomas_reliance_2022}, argue that we could not.

The predominant evaluation practice of relying on automatic scores is questioned in many contexts in NLP today, especially in NLG \citep{callison-burch-etal-2006-evaluating,liu-etal-2016-evaluate,novikova-etal-2017-need,sulem-etal-2018-bleu,reiter-2018-structured}.
In the context of explainability, the need for \textit{human-centered} evaluation is stressed by, i.a., \cite{DBLP:conf/iui/RiberaL19}, \cite{DBLP:journals/corr/abs-2007-12248}, \cite{DBLP:conf/acl/GonzalezRS21}, \cite{https://doi.org/10.48550/arxiv.2112.04417}, \cite{DBLP:conf/icwsm/SchlegelGB22} or \cite{liao2022connecting}.
We argue that human evaluation always has to be part of explanation quality evaluation.
User studies yield insights beyond proxy scores as they can comprise (i) a broader set of quantifiable dimensions than proxy scores can offer as well as (ii) dimensions of explanation quality that are inaccessible using quantitative methods at all but require \textit{qualitative} approaches, such as mental model analysis \citep{DBLP:journals/corr/abs-2002-02526, kulesza_too_2013} or thematic analysis \citep{braun2006using} in which themes are extracted from textual responses or transcriptions via various steps (coding, theme generation and review etc.).

We argue that searching for valuable systems based on proxy metrics alone can be regarded to be an instance of the \textbf{\textit{Streetlight Effect}}, also know as the \textit{Drunkard's Search Principle} \citep{kaplan_conduct_1964,iyengar_explorations_1993}.
This effect describes a situation in which a drunken man lost his keys in a park, but instead of searching for them in the place where he lost them, he is \textbf{searching under a streetlight because this is where the light is}. 
We argue that we face a similar situation when we exclusively rely on proxy metrics as shown in \Cref{fig:streetlight_effect}.
Instead of focusing on what we ultimately are interested in, i.e., providing good explanations to users, we narrow our focus onto increasing proxy scores instead.
To shed light on a broader spectrum of explanation quality, our quantitative measures should include both, validated proxy scores and human ratings/signals.

\paragraph{Case Study}
Our results reported in \Cref{sec:p_unexplored_proxy_scores} already show that human ratings exceed the information we are able to get from our investigated proxy scores.
In addition, there also is information that we cannot obtain using human ratings alone.
To illustrate this, we collect voluntary free text feedback from our participants:

\begin{itemize}
	\item ``\textit{I see why the model thought it, but it doesn't provide any useful info in reality}''. \\
 This comment shows that users have the impression that a model ``thinks'', hinting at anthropomorphisation.
 Concretely, this suggests to consider the inclusion of an anthropomormphism questionnaire in subsequent user studies. 
	\item ``\textit{The question asks about two players but there is only  a correct answer for one player and only one explanation}''.\\
 This comment confirms that one type of model error is to provide answers that do not semantically match the question. Consequently, developing a new proxy score to quantify the semantic overlap  between the predicted answer and the question could help to guide model development. 
	\item ``\textit{It doesn't really state how it came up with this answer, as it only told about other fights. My default answer is incorrect, until the system proves it to be true.}''\\
 This comment informs us about the user's rating behavior and suggests that a re-wording of the question could allow us to capture the range of perceived correctness in a better way.
\end{itemize}

\vspace{0.2cm}
\casestudybox{
Overall, the collected comments illustrate that \textbf{qualitative evaluation can yield insights beyond quantitative participant ratings} which in turn can help to improve proxy scores and human rating evaluation.
}

\subsection{Single-score Leaderboards}\label{sec:p_single_metric}
The current practice in NLP leaderboards (and many NLP research work in general) is the scoring and comparing of systems using a single score such as accuracy, BLEU or F1.
In \Cref{sec:c_multidimensional}, we already motivated that explanation quality has multiple independent dimensions. Therefore, it should be measured with multiple scores. Moreover, aggregating those scores (e.g., via averaging) to obtain a single measure will not be expedient either since the dimensions might be independently useful and/or scaled differently.

Ranking systems using a single score can also lead to over-optimization of this one score \citep{thomas_reliance_2022} and can lead to the deterioration of score validity as we argued and demonstrated in \Cref{sec:goodhart}.
This could be prevented by using a diverse set of scores instead of only one score.

\begin{table}%[t]
    \centering
    \resizebox{0.99\textwidth}{!}{%
    \begin{tabular}{p{0.01\textwidth} C{0.18\textwidth} C{0.18\textwidth} C{0.18\textwidth} C{0.18\textwidth} C{0.18\textwidth}}
    \toprule
        & \multicolumn{5}{c}{\textbf{ranking criterion}}\\
        \cmidrule(rl){2-5}\cmidrule(rl){6-6}
        & \textbf{joint-F1} & \textbf{LocA} & \textbf{averaged proxy scores} & \textbf{factor-weighted proxy scores}  & \textbf{human usability ratings}\\
        \cmidrule(rl){2-5}\cmidrule(rl){6-6}
        1 &\namewithnumber{\textit{gold}}{1.00} &  \namewithnumber{\textit{gold}}{1.00}               & \namewithnumber{\textit{gold}}{1.00}                        & \namewithnumber{\textit{gold}}{1.73}  & \namewithnumber{FE2H on ALBERT}{0.98} \\
         \cmidrule(rl){2-5}\cmidrule(rl){6-6}
        2 & \namewithnumber{FE2H on ALBERT}{0.77}                                                              &  \namewithnumber{gold-answers-all-facts}{1.00}      & \namewithnumber{FE2H on ALBERT}{0.78}                & \namewithnumber{FE2H on ALBERT}{1.47} & \namewithnumber{HGN}{0.90}\\
         \cmidrule(rl){2-5}\cmidrule(rl){6-6}
        3 & \namewithnumber{AMGN}{0.74}                                                                     &  \namewithnumber{FE2H on ALBERT}{0.98}                 & \namewithnumber{AMGN}{0.76}                       & \namewithnumber{AMGN}{1.43} & \namewithnumber{\fbox{S2G-large}}{0.88}\\
         \cmidrule(rl){2-5}\cmidrule(rl){6-6}
        4 & \namewithnumber{Longformer}{0.73}                                                               &  \namewithnumber{HGN}{0.97}                         & \namewithnumber{HGN}{0.74}                        & \namewithnumber{HGN}{1.4}  & \namewithnumber{Longformer}{0.87}\\
         \cmidrule(rl){2-5}\cmidrule(rl){6-6}
        5 & \namewithnumber{\fbox{S2G-large}}{0.72}                                                                &  \namewithnumber{AMGN}{0.95}                        & \namewithnumber{Longformer}{0.74}                 & \namewithnumber{Text-CAN}{1.31}  & \namewithnumber{SAE}{0.87}\\
         \cmidrule(rl){2-5}\cmidrule(rl){6-6}
        6 & \namewithnumber{HGN}{0.71}                                                                      &  \namewithnumber{Text-CAN}{0.92}                    & \namewithnumber{Text-CAN}{0.70}                   & \namewithnumber{Longformer}{1.28}  & \namewithnumber{Text-CAN}{0.87}\\
         \cmidrule(rl){2-5}\cmidrule(rl){6-6}
        7 & \namewithnumber{Text-CAN}{0.66}                                                                 &  \namewithnumber{GRN}{0.89}                         & \namewithnumber{\fbox{S2G-large}}{0.69}                  & \namewithnumber{SAE}{1.25}   & \namewithnumber{AMGN}{0.87}\\
         \cmidrule(rl){2-5}\cmidrule(rl){6-6}
        8 & \namewithnumber{SAE}{0.63}                                                                      &  \namewithnumber{SAE}{0.86}                         & \namewithnumber{SAE}{0.67}                        & \namewithnumber{gold-answers-all-facts}{1.23} & \namewithnumber{gold-answers-all-facts}{0.86}\\
         \cmidrule(rl){2-5}\cmidrule(rl){6-6}
        9 & \namewithnumber{IRC}{0.59}                                                                      &  \namewithnumber{IRC}{0.77}                         & \namewithnumber{GRN}{0.64}                        & \namewithnumber{GRN}{1.21} & \namewithnumber{\textit{gold}}{0.83}\\
         \cmidrule(rl){2-5}\cmidrule(rl){6-6}
        10 & \namewithnumber{GRN}{0.58}                                                                      &  \namewithnumber{Longformer}{0.72}                  & \namewithnumber{IRC}{0.62}                       & \namewithnumber{IRC}{1.17} & \namewithnumber{IRC}{0.83}\\
         \cmidrule(rl){2-5}\cmidrule(rl){6-6}
        11 & \namewithnumber{gold-answers-all-facts}{0.12}                                                   &  \namewithnumber{random-answers-gold-facts}{0.12}   & \namewithnumber{gold-answers-all-facts}{0.53}    & \namewithnumber{\fbox{S2G-large}}{0.94} & \namewithnumber{GRN}{0.68}\\
         \cmidrule(rl){2-5}\cmidrule(rl){6-6}
        12 & \namewithnumber{random-answers-all-facts}{0.02}                                                 &  \namewithnumber{\fbox{S2G-large}}{0.12}                   & \namewithnumber{random-answers-gold-facts}{0.3}  & \namewithnumber{random-answers-gold-facts}{0.09} & \namewithnumber{random-answers-random-facts}{0.23}\\
         \cmidrule(rl){2-5}\cmidrule(rl){6-6}
        13 & \namewithnumber{gold-answers-random-facts}{0.00}                                                &  \namewithnumber{random-answers-random-facts}{0.11} & \namewithnumber{gold-answers-random-facts}{0.29} & \namewithnumber{random-answers-random-facts}{0.06} & \namewithnumber{random-answers-gold-facts}{0.21}\\
         \cmidrule(rl){2-5}\cmidrule(rl){6-6}
        14 & \namewithnumber{random-answers-random-facts}{0.00}                                              &  \namewithnumber{gold-answers-random-facts}{0.03}   & \namewithnumber{random-answers-random-facts}{0.01}& \namewithnumber{gold-answers-random-facts}{0.02} & \namewithnumber{gold-answers-random-facts}{0.16}\\
        \bottomrule
    \end{tabular}%
    }
    \vspace{0.1cm}
    \caption{Ranking models with respect to different criteria. We construct leaderboards for (a) joint-F1 (official leaderboard score), (b) the answer-explanation consistency measure LocA, (c) the average over all 14 proxy scores, (c) a factor-loading-weighted average over the three proxy scores which we found to be associated with human ratings within our factor analysis, and (d) human utility ratings. We mark \fbox{S2G-large} and \textit{gold predictions} to demonstrate inconsistent model ranks across criteria.
    }
    \label{tab:different_leaderboards}
\end{table}

\paragraph{Case Study}
We construct various leaderboards from the HotpotQA systems and evaluate how sensitive model rankings are with respect to the ranking criterion.
\Cref{tab:different_leaderboards} displays the respective model rankings of four criteria.

We observe that \textbf{different scores and weighting schemes lead to contradicting model rankings}.
For example, S2G-large (marked with a box in \Cref{tab:different_leaderboards}) is rated the fourth-best real model according to joint-F1 and the third-best regarding usability ratings but rated the worst real model according to LocA.
While all real models except FE2H on ALBERT differ with respect to their relative rankings, FE2H on ALBERT is ranked as the best real model across all criteria including human usability ratings, indicating that this model offers substantial benefits over the other models.

Further, \textbf{rankings regarding human ratings and proxy metrics disagree heavily} as we can see for the gold predictions that consistently are ranked top following the proxy score leaderboards, but are ranked eighth following human usability ratings.
Interestingly, the gold answers along with \textit{all facts} are ranked as more usable then the gold facts with only the relevant facts.
\vspace{0.2cm}
\casestudybox{
Overall, our results signal a disagreement between the user needs \textit{assumed} within the HotpotQA dataset and the \textit{actual} user needs within our participant sample. 
}

\section{Guidelines}\label{sec:guidelines}
This section proposes guidelines to address the shortcomings described in \Cref{sec:problems}.

\subsection{Validate Proxy Scores Against Humans}\label{sec:s_relation_proxy_human}
While there is a lot of work on investigating the relation between automatic scores and human ratings in  natural language generation  \citep{belz-reiter-2006-comparing,novikova-etal-2017-need,dusek-etal-2019-automatic}, only few studies consider this aspect in the context of explanation evaluation \citep{jannach_escaping_2020,schuff-etal-2020-f1,schuff-etal-2021-external,DBLP:conf/iccv/KayserCSEDAL21,clinciu-etal-2021-study}.
To address the problem of unvalidated proxy scores for explanation quality evaluation (\Cref{sec:p_unexplored_proxy_scores}), we advise to consistently validate the relation between proxy scores and human signals, such as human-AI performance, subjective ratings, completion times or physiological measures like eye tracking.
One straight-forward approach to quantify these relations is a correlation analysis.

\paragraph{Advantages}
Given proxy scores that yield sufficiently strong correlations to human ratings/signals, those scores can be used to develop systems that are actually useful for users.

\paragraph{Limitations}
Given a new task or leaderboard, it is unlikely that we have access to a representable pool of models which can be used to validate the metrics. Therefore, we have to accept a certain \textbf{grace period} in which we can only assume that the chosen evaluation scores lead to reasonable results. Once there is a handful of models available, the proxy metrics should then be validated against human scores and revised if necessary.

Referring to our discussion of Goodhart's law in \Cref{sec:p_single_metric}, any proxy metric has to be \textbf{periodically re-tested} for its validity.\footnote{The need for re-testing can be recognized by, e.g., monitoring demographic changes in the target population and/or changes in the correlations within user ratings.
}

Finally, each validity evaluation is limited to a \textbf{group of explainees} (see \Cref{sec:c_explainees}). Different groups of users will have different needs and, as a result, explanation quality evaluation will need different measures. For example, validity findings for a population of high-school students might not transfer to NLP researchers.

\subsection{Do Human Evaluation}\label{sec:s_user_studies}
In \Cref{sec:s_relation_proxy_human}, we already recommend user studies for the purpose of proxy score validation.
Based on our discussion in \Cref{sec:p_missing_user_studies}, we also propose to do as much human evaluation as possible in order to gain \textbf{additional explanation quality indicators} from human rating scores directly.
In the context of application-oriented model development, human evaluation can be conducted as the final evaluation step after model tuning.
In the context of leaderboards, we propose to regularly conduct human assessments of (a subset) of system submissions.

\paragraph{Measures of Human Behavior and Perceived Quality}
When choosing what to measure within a user study, we suggest to \textbf{collect objective measures of user behavior as well as subjective ratings}.
\Cref{tab:human_signals} lists a selection of possible measures of (a) objective measures of human behavior (top) and (b) subjective human ratings (bottom) along with exemplary publications including the respective scores/ratings.\footnote{We refer to \cite{DBLP:conf/iui/ChromikS20} as well as \cite{https://doi.org/10.48550/arxiv.2201.08164} for a review of (quantitative) human evaluation methods conducted in explainability research.}

\begin{table}[t]
    \centering
    \resizebox{0.9\textwidth}{!}{%
    \begin{tabular}{C{0.03cm}C{0.22\columnwidth}L{0.405\columnwidth}L{0.25\columnwidth}}
        \toprule
          & \textbf{Measure} & \textbf{Description} & \textbf{References}  \\
         \midrule
         \multirow{18}{*}{\rotatebox[origin=c]{90}{\textit{objective scores}}} & Time & Time measures of, e.g., task completion or interaction with the system & \cite{DBLP:conf/chi/LimDA09}, \cite{Lage-Chen-He-Narayanan-Kim-Gershman-Doshi_Velez-2019}, \cite{DBLP:conf/chi/ChengWZOGHZ19}, \cite{schuff-etal-2020-f1}\\
         & Human performance & Task performance of users (e.g., accuracy in an AI-supported decision task) & \cite{DBLP:conf/iui/FengB19}, \cite{schuff-etal-2020-f1}, \cite{lage2019human}, \cite{bansal_does_2021}\\
         & Simulatability & Measures related to how well explanations enable users to predict system performance (given an explanation when making predictions on new instances) & \cite{hase-bansal-2020-evaluating}, \cite{DBLP:conf/iui/WangY21}\\
         & Teachability & Measures related to how well explanations enable users to predict system performance (without having access to explanations when making predictions on new instances)& \cite{DBLP:conf/icml/GoyalWEBPL19}, \cite{DBLP:conf/cvpr/WangV20}\\
         & Agreement & Frequency of how often a user accepts a system decision & \cite{DBLP:conf/fat/ZhangLB20}, \cite{bansal_does_2021}\\
         & Number of user interactions & Number of times a user, e.g., runs a model to predict an output & \cite{pezeshkpour-etal-2022-combining} \\
         \cmidrule(lr){1-4}
         \multirow{16}{*}{\rotatebox[origin=c]{90}{\textit{subjective ratings}}} & Perceived performance & Subjective estimate of system performance & \cite{nourani2019effects}, \cite{schuff-etal-2020-f1}\\
         & Over- / underestimation & Difference between perceived system performance and actual system performance & \cite{Nourani-Kabir-Mohseni-Ragan-2019}, \cite{schuff-etal-2020-f1} \\
         & Trust & Trust in the model's abilities/correctness & \cite{bussone2015role}, \cite{schuff-etal-2020-f1}, \cite{DBLP:conf/interact/RibesHPDPGS21}, \cite{DBLP:journals/pacmhci/BucincaMG21} \\
         & Perceived usefulness & User-reported system usefulness & \cite{DBLP:conf/vl/KhuranaAC21}, \cite{bansal_does_2021} \\
         & Subjective understanding & Self-reported degree of system understanding & \cite{DBLP:conf/iui/EhsanTCHR19}, \cite{DBLP:conf/iui/WangY21}, \cite{DBLP:conf/interact/RibesHPDPGS21} \\
         & Grammaticality & Ratings or grammatical correctness & \cite{schuff-etal-2021-external}, \cite{liu-etal-2022-generated} \\
         & Perceived factuality & Ratings of factual correctness & \cite{schuff-etal-2021-external}, \cite{liu-etal-2022-generated} \\
         & Mental demand & Self-reports of mental demand in processing the explanation & \cite{DBLP:journals/pacmhci/BucincaMG21}\\
         \bottomrule
    \end{tabular}
    }
    \vspace{0.3cm}
    \caption{Selection of (a) scores of objective human behavior (top) and (b) dimensions of subjective self-reports of perceived quality (bottom).}
    \label{tab:human_signals}
\end{table}

Objective measures include, e.g., response time, human task performance, human-AI agreement, but also more complex scores, e.g., Utility-$k$ \citep{https://doi.org/10.48550/arxiv.2112.04417}.

Subjective ratings include, e.g., perceived accuracy, trust, perceived usefulness \citep{liu-etal-2022-generated}, or mental demand \citep{DBLP:journals/pacmhci/BucincaMG21}.
As noted by \cite{DBLP:conf/iui/BucincaLGG20}, subjective ratings should complement objective measures of user performance as the latter cannot necessarily be inferred from the former \citep{DBLP:conf/iui/BucincaLGG20}.

Following \citet{jannach_escaping_2020} and \citet{thomas_reliance_2022}, we further advocate to also \textbf{collect qualitative feedback} (e.g., participant comments within a user study or a focus group) to complement quantitative measures.
We demonstrate how qualitative feedback can yield insights beyond quantitative evaluation within our case study in \Cref{sec:p_missing_user_studies}.

The study conducted by \cite{DBLP:conf/chi/ChengWZOGHZ19} is a good example of how objective measures can be combined with both qualitative as well as quantitative human evaluation.
Additional examples of such \textit{mixed-methods} evaluations can be found in the work of \cite{bansal_does_2021} in the context of complementary human-AI team performance and \cite{https://doi.org/10.48550/arxiv.2302.00096} in the context of clinical AI acceptance.
Note, however, that collecting qualitative feedback via, e.g., the think-aloud method, can impact users' mental effort allocation and potentially impact participant behavior \citep{DBLP:conf/iui/BucincaLGG20} and, consequently, studies should be designed carefully.

\paragraph{Advantages}
Human evaluation allow us to re-adjust the direction into which we develop systems by unveiling explanation quality dimensions that were previously hidden.
For example, qualitative findings from user comments can help us to identify system qualities we did not think of before.
Moreover, human evaluations can reward the development of systems that follow an unconventional approach and, as a result, whose explanation qualities might be undetectable using proxy scores.
This can motivate researchers to develop original models and can ultimately \textbf{diversify and accelerate} research.

\paragraph{Limitations}
Each human evaluation is bound to noise w.r.t. the pool of participants and the way they approach the study (for example whether they carefully read the questions).
However, in contrast to \emph{annotation} (on an instance level), noisy human responses do not have to limit human \textit{evaluation} (on a system level) using adequate statistical tools.

Further, potentially high costs to compensate the participants and longer preparation times to recruit participants and conduct and carefully evaluate the studies might hinder the conduction of a user study.

Additionally, proxy task evaluations (i.e., evaluations that are conducted on simplified human-AI tasks) do not necessarily lead to the same findings that real human-AI tasks yield and, in fact, can even contradict the latter \citep{DBLP:conf/iui/BucincaLGG20}.

Finally, user study interfaces have to be designed carefully as presumably minor design choices can heavily affect participant behavior.
For example, \cite{sullivan-etal-2022-explaining} find participants' rational explanation selections (i.e., marking relevant words in the text input) was greatly impacted by whether participants could mark multiple words at once or not.
Introductory texts on designing and conducting user studies in NLP can be found in, e.g., \cite{belz-etal-2020-disentangling} (NLG), \cite{iskender-etal-2021-reliability} (text summarization), \cite{sedoc-etal-2019-chateval} (chat bots), or \cite{schuff_vanderlyn_adel_vu_2023} (general NLP).

\paragraph{Case Study}
We discuss the experiment design of our case study along with a description of collected ratings in \Cref{sec:case_study_human_evaluation}.
A detailed discussion of our results can be found in the ``case study'' paragraphs across \Cref{sec:problems}.

\vspace{0.2cm}
\casestudybox{
Overall, our human evaluation allowed us to identify low correlations between human ratings and proxy scores, detect that correlations decreased over three years of system submissions, qualitative user feedback helps to spot shortcomings of proxy scores and human rating evaluation, and systems ranks based on proxy scores conflict with system ranks based on human ratings.
}

\subsection{Report Various Scores Without Averaging}
As we argued in \Cref{sec:p_single_metric}, using a single score for evaluation (regardless of proxy scores or human ratings/signals) can be misleading.
Therefore, we propose to use various scores rather than weighting quality dimensions against each other to get a single score.
This is in line with the recommendations by \citet{thomas_reliance_2022}.
While prior work proposed alternative leaderboards using on-demand (crowdsourcing) evaluation \citep{chaganty-etal-2017-importance} and personalized utility rankings \citep{ethayarajh-jurafsky-2020-utility}, we are---to the best of our knowledge---the first to provide a \textbf{leaderboard that does not condense multiple scores into a single one}.

\begin{figure}[t!]
    \centering
    \includegraphics[width=0.55\columnwidth]{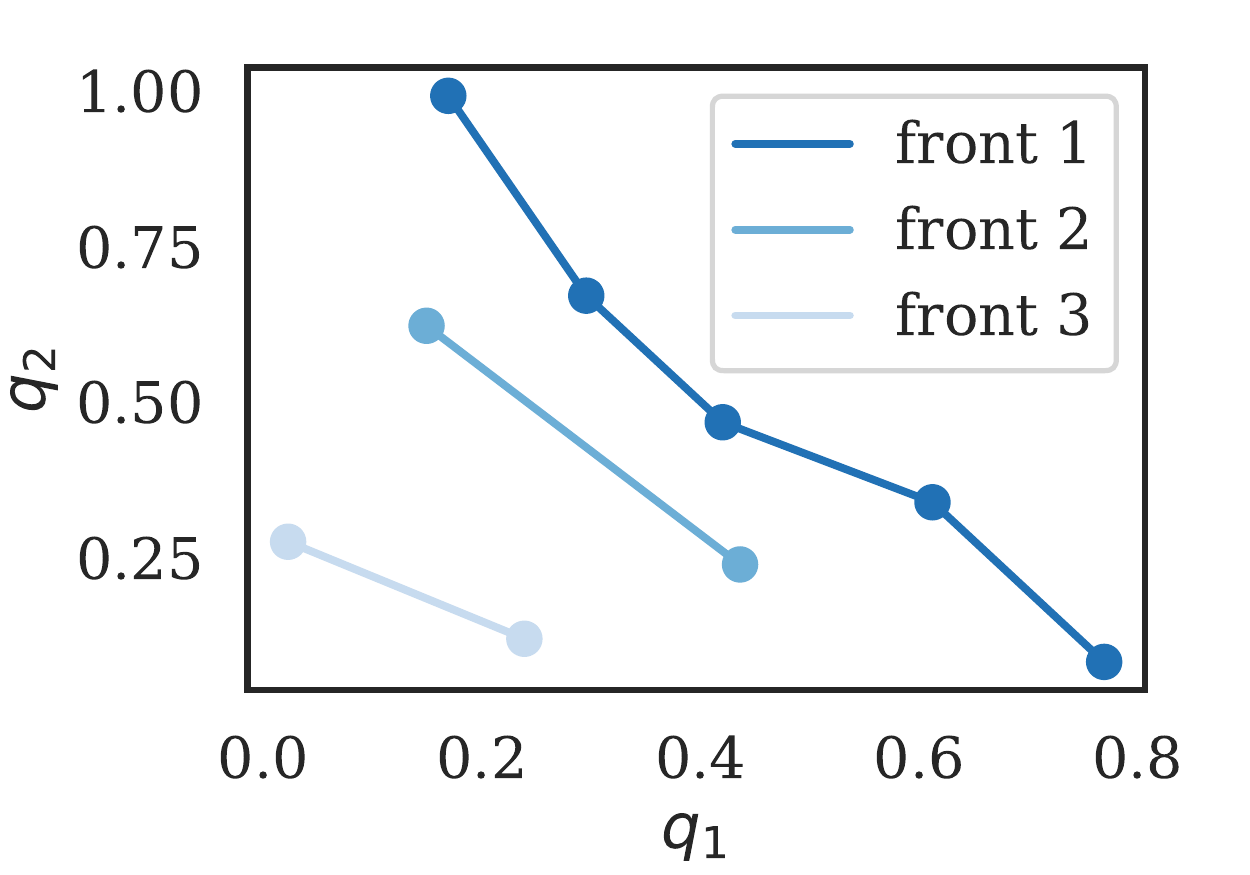}
    \vspace{0.3cm}
    \caption{Ranked Pareto fronts for two dimensions and 9 (fictional) systems. Each point represents a system along two (higher-is-better) scores $q_1$ and $q_2$.
    }
    \label{fig:pareto_front_example}
\end{figure}

\paragraph{Pareto Front Leaderboards}
To compare systems based on multiple scores, e.g., on a leaderboard, we propose to leverage the concept of \emph{Pareto efficiency}.
In the context of multidimensional leaderboards, a system is called Pareto efficient if the only way to select another system that is better regarding any score dimension is to worsen another score dimension.
For example, system A is Pareto efficient if the only way to select another system to increase, e.g., the F1 score, is to choose a system that has a lower, e.g., accuracy.
Given a set of systems, multiple systems can simultaneously be Pareto efficient.
\Cref{fig:pareto_front_example} shows a fictional example with nine systems (visualized by points) and two higher-is-better quality scores $q_1$ and $q_2$ (visualized by axes).
All five systems on the so-called Pareto front (front 1) are Pareto efficient and thus have rank 1.
To rank the remaining systems, we remove those five systems, calculate the next Pareto front (front 2), and repeat this until all systems are ranked.
The resulting leaderboard of the example shown in \Cref{fig:pareto_front_example} would consequently have five models on the first place (i.e., front), two models on the second and two models on the third.

\paragraph{Related Applications of Pareto Efficiency}
We are not the first to leverage Pareto efficiency within NLP.
\cite{pimentel-etal-2020-pareto} use Pareto efficiency to propose a new probing approach that trades-off probe accuracy and complexity.
In contrast to their work, we use Pareto efficiency to construct leaderboards.
Similar to our approach, \cite{liu-etal-2022-towards-efficient} argue that, in the context of efficient NLP models, models should be judged in terms of how far they overstep the performance-efficiency Pareto front.
In contrast to their work, we do not only consider the (first) Pareto front, but extend the concept of Pareto-efficiency to multiple fronts which form the ranks of our proposed leaderboard.

\paragraph{Advantages}
Using multiple scores for evaluation offers the advantage of \emph{capturing diverse aspects} of a system.
If a sufficiently diverse set of scores is used, the \textbf{over-optimization of one score can be prevented} since other scores would likely be decreased at the same time.
This is supported by \textit{surrogation} effects \citep{choi_lost_2012,choi_strategy_2013} where, in the context of manager compensation, \citet{choi_lost_2012} find that  manager decisions can be improved when ``managers are compensated on multiple measures of a strategic construct'' instead of on a single one.
We hypothesize that this observation also holds for AI practitioners that need to choose a system, e.g., from a leaderboard.

When using Pareto front leaderboards, we can \textbf{rank systems without weighting the different quality dimensions against each other}. In particular, the concept of Pareto efficiency allows us to choose systems that are \emph{not worse} than others on all fronts.
Note that Pareto fronts are robust to score re-scaling and are applicable to ordinal (e.g., Likert) ratings.

\paragraph{Limitations}
With multiple scores, it can be hard to determine a ``winning'' system because \emph{different models might rank best} on different scores.  
Pareto Front Leaderboards can mitigate this problem, however, they may result in a set of winning systems instead of a single winning system.
We argue that this is not a real limitation though since the concept of Pareto efficiency ensures that a system on one front is not worse than other systems on the same front.

In the extreme case when the number of scores is high in comparison to the number of systems that should be scored, the resulting leaderboard can collapse to a single front because the fronts' surface grows exponentially with the number of scores.
We therefore recommend to ensure that the number of variables should only be increased along with a sufficient increase in the number of systems.

Further, Pareto Front leaderboards can be ``attacked'' by optimizing a single metric with the purpose of positioning a new system inside the first front.
Although this allows the leaderboards to be gamed to a certain extent, a truly remarkable improvement is one that creates a new front which is, in turn, robust to the improvement of single metrics.

\begin{table}
    \centering
    \resizebox{0.99\textwidth}{!}{%
    \begin{tabular}{C{0.07\textwidth}L{0.8\textwidth}}
    \toprule
     \textbf{Rank} & \textbf{Models} (original HotpotQA ranks in parantheses) \\
    \midrule
        1 & gold (*), random-answers-gold-facts (*), FE2H on ALBERT (3), Longformer (25), S2G-large (31), HGN (35), Text-CAN (47), IRC (63) \\
        2 & AMGN (16), SAE (48), GRN (65),
        DecompRC (unranked), random-answers-random-facts (*), gold-answers-all-facts (*)\\
        3 &                                        gold-answers-random-facts (*) \\
    \bottomrule
    \end{tabular}%
    }
    \vspace*{0.1cm}
    \caption{Ranked Pareto fronts based on \emph{human ratings}. ``$\ast$'' marks systems derived from the ground truth annotations.
    }\label{tab:human_ranked_pareto_fronts}
\end{table}

\begin{table}
    \centering
    \resizebox{0.99\textwidth}{!}{%
    \begin{tabular}{C{0.07\textwidth}L{0.87\textwidth}}
    \toprule
     \textbf{Rank} & \textbf{Models} (original HotpotQA ranks in parantheses)\\
    \midrule
        1 &                                                                 gold (*) \\
        2 & gold-answers-all-facts (*), rand.-answers-gold-facts (*), FE2H on ALBERT (3), AMGN (16) \\
        3 &      Longformer (25), HGN (35), IRC (63), gold-answers-random-facts (*) \\
        4 &                                                  S2G-large (31), Text-CAN (47) \\
        5 &                                                           SAE (48),  GRN (65) \\
        6 &                                                         DecompRC (unranked) \\
        7 &                                          random-answers-random-facts (*) \\
    \bottomrule
    \end{tabular}%
    }
    \vspace*{0.1cm}
    \caption{Ranked Pareto fronts based on \emph{proxy scores}.
    }\label{tab:automatic_ranked_pareto_fronts}
\end{table}

\paragraph{Case Study}
We evaluate the 15 models described in \Cref{sec:task_model_evaluation} on numerous (i) human ratings and (ii) automatic scores.
Then, we construct two Pareto front leaderboards, one for human ratings and one for automatic scores.

\Cref{tab:human_ranked_pareto_fronts} shows the Pareto front leaderboard based on human ratings (usability, mental effort, utility, correctness, consistency and completion time).
We observe that high-performing models, such as
FE2H on ALBERT (official leaderboard rank 3) are located within the rank 1 Pareto front en-par with the gold prediction system.
Interestingly, previously lower-ranked models, such as IRC (leaderboard rank 63) are also located in the first Pareto front which means that they also possess a combination of strengths that dominates the models in the other ranks.

\Cref{tab:automatic_ranked_pareto_fronts} shows the leaderboard based on automatic proxy scores.
The gold prediction system is the single winner in this leaderboard, followed by the two real models FE2H on ALBERT and AMGN.
While the first models are ordered consistently with the HotpotQA leaderboard, the Pareto front leaderboards disagrees w.r.t. ranks for others, e.g., the IRC model (leaderboard rank 63), Longformer (leaderboard rank 25) or S2G-large  (leaderboard rank 31).
For the synthetic systems, we observe differences across the two Pareto front leaderboards.
For example, the gold-answers-random-facts system is ranked last w.r.t. human ratings but ranked third w.r.t. automatic scores.

\vspace{0.2cm}
\casestudybox{
Our results highlight, again, that proxy metrics do not reflect the quality dimensions probed in the human ratings sufficiently well.
We provide details on the exact proxy scores and model ratings in \Cref{sec:detailed_proxy_scores} and \Cref{sec:human_rating_details}.
}

\section{Conclusion}
This paper aims at increasing the awareness of the shortcomings and open challenges that today's explanation quality evaluation practices face.
We discuss general characteristics of explanation quality, describe current practices and point out to which extent they violate the discussed characteristics.
We support our arguments with empirical evidence of a crowdsourced case study that we conducted for the example of explainable question answering systems from the HotpotQA leaderboard.

Concretely, we demonstrate that (i) proxy scores poorly reflect human explanation quality ratings, (ii) proxy scores can loose their expressiveness over time, (iii) human evaluation yields quantitative as well as qualitative insight beyond automatic evaluation, and (iv) single-score leaderboards fail to reflect the spectrum of explanation quality dimensions. 

We propose (a) guidelines for a more effective and human-centered evaluation as well as (b) an alternative type of leaderboard that constructs ranks from multiple dimensions without averaging scores.
We aim to inform and inspire future work and ultimately drive the field towards reliable and meaningful explanation quality evaluation. 

\backmatter
\bmhead{Acknowledgments}
We are grateful to Annemarie Friedrich for valuable feedback and helpful discussions.
We also thank all HotpotQA system submittors that allowed us to include their systems in our analysis.

\clearpage
\begin{appendices}

\appendix 

\section{Detailed Proxy Scores}\label{sec:detailed_proxy_scores}
We provide values of all analyzed proxy scores for all analyzed  10 leaderboard systems as well as our five synthetic systems in \Cref{tab:leaderboard}.

\FloatBarrier

\begin{table}
    \centering
    \resizebox{\textwidth}{!}{
    \begin{tabular}{lrrrrrrr}
        \toprule
        {} &  joint\_f1 &      f1 &   sp\_f1 &  \_loca\_score &  num\_words &  num\_facts &  num\_excess\_facts \\
        \midrule
        gold                        &    0.9999 &  0.9999 &  1.0000 &       0.9980 &    58.4296 &     2.4305 &            0.0000 \\
        fe2h\_albert                 &    0.7654 &  0.8444 &  0.8914 &       0.9823 &    56.0451 &     2.3020 &           -0.1286 \\
        amgn\_plus                   &    0.7525 &  0.8338 &  0.8883 &       0.9596 &    56.4771 &     2.3291 &           -0.1014 \\
        fe2h\_electra                &    0.7491 &  0.8269 &  0.8872 &       0.9860 &    55.7048 &     2.2872 &           -0.1433 \\
        s2g\_plus                    &    0.7436 &  0.8217 &  0.8873 &       0.1261 &    55.5947 &     2.2758 &           -0.1548 \\
        hgn\_large                   &    0.7422 &  0.8220 &  0.8848 &       0.9719 &    55.9020 &     2.3045 &           -0.1260 \\
        amgn                        &    0.7420 &  0.8280 &  0.8812 &       0.9547 &    54.0543 &     2.2154 &           -0.2151 \\
        git                         &    0.7387 &  0.8201 &  0.8819 &       0.9691 &    56.7310 &     2.3357 &           -0.0948 \\
        ffreader\_large              &    0.7379 &  0.8216 &  0.8843 &       0.9203 &    56.3926 &     2.3213 &           -0.1070 \\
        longformer                  &    0.7317 &  0.8126 &  0.8834 &       0.7154 &    56.4994 &     2.3255 &           -0.1051 \\
        hgn\_roberta\_large           &    0.7258 &  0.8062 &  0.8761 &       0.9829 &    57.2278 &     2.3661 &           -0.0644 \\
        text\_can\_large              &    0.7253 &  0.8080 &  0.8696 &       0.9519 &    56.1457 &     2.3049 &           -0.1256 \\
        s2g\_large                   &    0.7226 &  0.8024 &  0.8761 &       0.1194 &    55.8983 &     2.3055 &           -0.1251 \\
        sae\_large                   &    0.7145 &  0.7963 &  0.8687 &       0.9583 &    56.2038 &     2.3067 &           -0.1238 \\
        hgn                         &    0.7104 &  0.7937 &  0.8733 &       0.9682 &    57.3562 &     2.3606 &           -0.0700 \\
        s2g\_base                    &    0.6952 &  0.7703 &  0.8720 &       0.1233 &    56.4389 &     2.3271 &           -0.1034 \\
        text\_can                    &    0.6596 &  0.7399 &  0.8576 &       0.9212 &    56.6486 &     2.3294 &           -0.1011 \\
        sae                         &    0.6292 &  0.7277 &  0.8282 &       0.8645 &    57.4972 &     2.3773 &           -0.0532 \\
        grn\_update                  &    0.6173 &  0.7023 &  0.8420 &       0.8493 &    52.1172 &     2.1249 &           -0.2971 \\
        mkgn                        &    0.6169 &  0.7069 &  0.8354 &       0.7238 &    54.2208 &     2.2308 &           -0.1997 \\
        qfe                         &    0.5962 &  0.6806 &  0.8450 &       0.8708 &    54.6789 &     2.2408 &           -0.1777 \\
        irc                         &    0.5922 &  0.7253 &  0.7936 &       0.7686 &    70.3375 &     2.9413 &            0.5107 \\
        grn                         &    0.5848 &  0.6672 &  0.8411 &       0.8886 &    57.3488 &     2.3715 &           -0.0470 \\
        kgnn                        &    0.5282 &  0.6575 &  0.7680 &       0.8608 &    55.6323 &     2.3449 &           -0.0856 \\
        decomprecomb                &    0.4517 &  0.6977 &  0.5938 &       0.1250 &    33.3648 &     1.3646 &           -1.0659 \\
        gold\_answers\_all\_facts      &    0.1179 &  0.9999 &  0.1180 &       0.9980 &   923.9561 &    41.2556 &           38.8251 \\
        random\_answers\_gold\_facts   &    0.0193 &  0.0193 &  1.0000 &       0.1205 &    58.4296 &     2.4305 &            0.0000 \\
        random\_answers\_random\_facts &    0.0000 &  0.0189 &  0.0000 &       0.1070 &    55.9535 &     2.4255 &           -0.0050 \\
        gold\_answers\_random\_facts   &    0.0000 &  0.9999 &  0.0000 &       0.0322 &    55.8294 &     2.4255 &           -0.0050 \\
        \bottomrule
        \end{tabular}
    }\vspace{0.1cm}
    \caption{Extended HotpotQA leaderboard including synthetic systems derived from the gold test set (marked with ``$\ast$'' and \textit{italics}). Best values are marked \textbf{bold} (no best values for \# words and \# facts). DecompRC is only evaluated on answer metrics.
    }
    \label{tab:leaderboard}
\end{table}

\section{User Study Details}
We show screenshots of our study interface in \Cref{sec:study_interface} and report human ratings along all measured rating dimensions in \Cref{sec:human_rating_details}.

\subsection{User Study Interface}\label{sec:study_interface}
\begin{figure*}
    \centering
    \includegraphics[width=\linewidth]{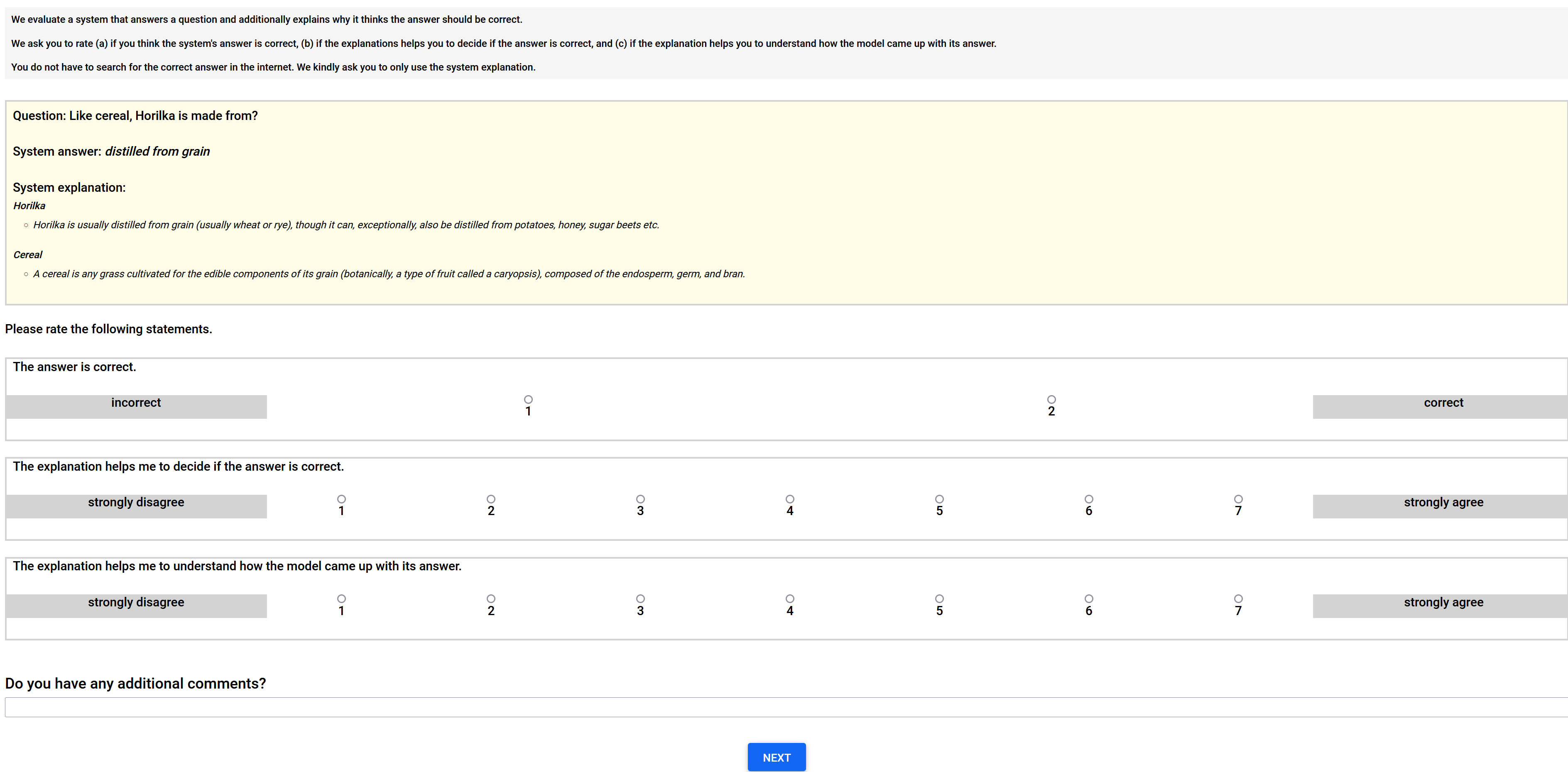}
    \caption{MTurk interface to rate a system prediction.}\label{fig:screenshot_instance}
\end{figure*}

\begin{figure*}
    \centering
    \includegraphics[width=\linewidth]{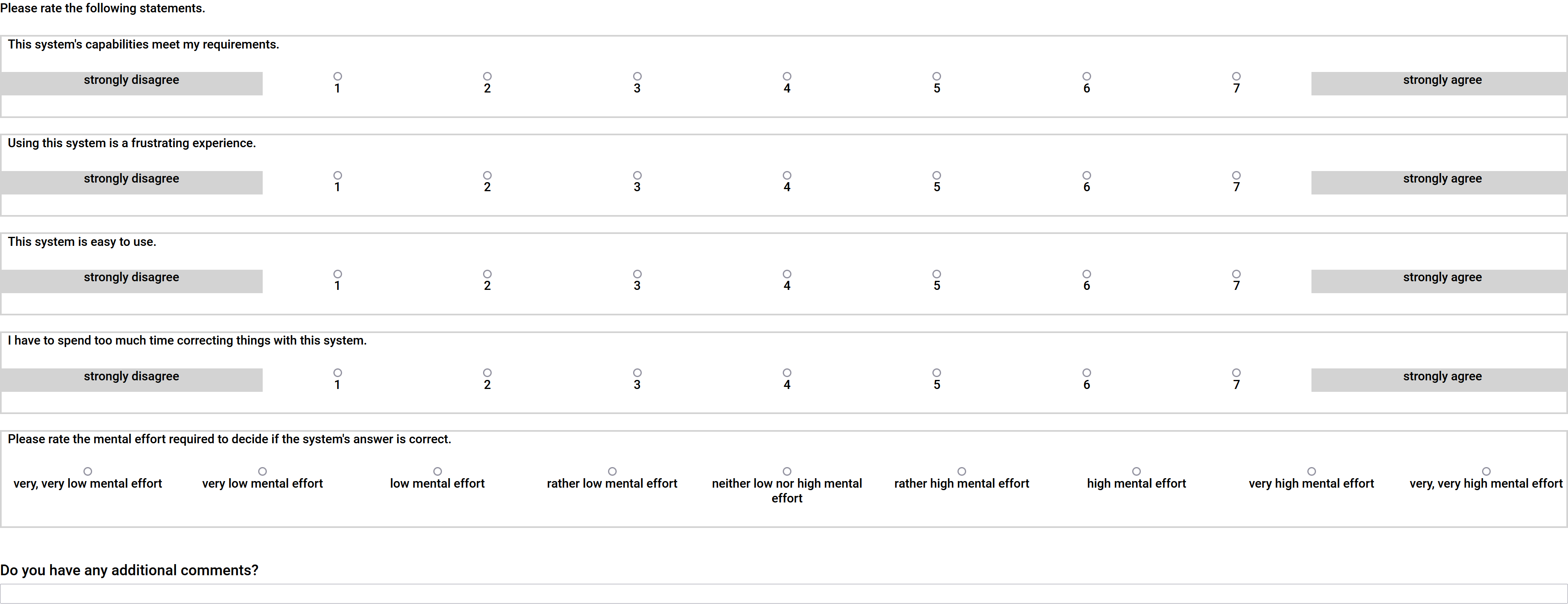}
    \caption{Post questionnaire of the MTurk interface.}\label{fig:screenshot_post_questionnaire}
\end{figure*}

We provide screenshots of the user study interface that we showed to participants.
\Cref{fig:screenshot_instance} displays the rating interface we showed \textit{for each question}.
\Cref{fig:screenshot_post_questionnaire} displays the post hoc questionnaire we asked participants to fill out at the end of the study (\textit{per system}, once per participant).

\subsection{Detailed Human Ratings}\label{sec:human_rating_details}
\Cref{tab:human_ratings} displays the human ratings and completion times we obtained within the user study for the 10 leaderboard systems as well as our five synthetic systems.

\subsection{Proxy Scores and Human Ratings}\label{automatic_scores_and_human_ratings}
\begin{figure*}
    \centering
    \includegraphics[width=\linewidth]{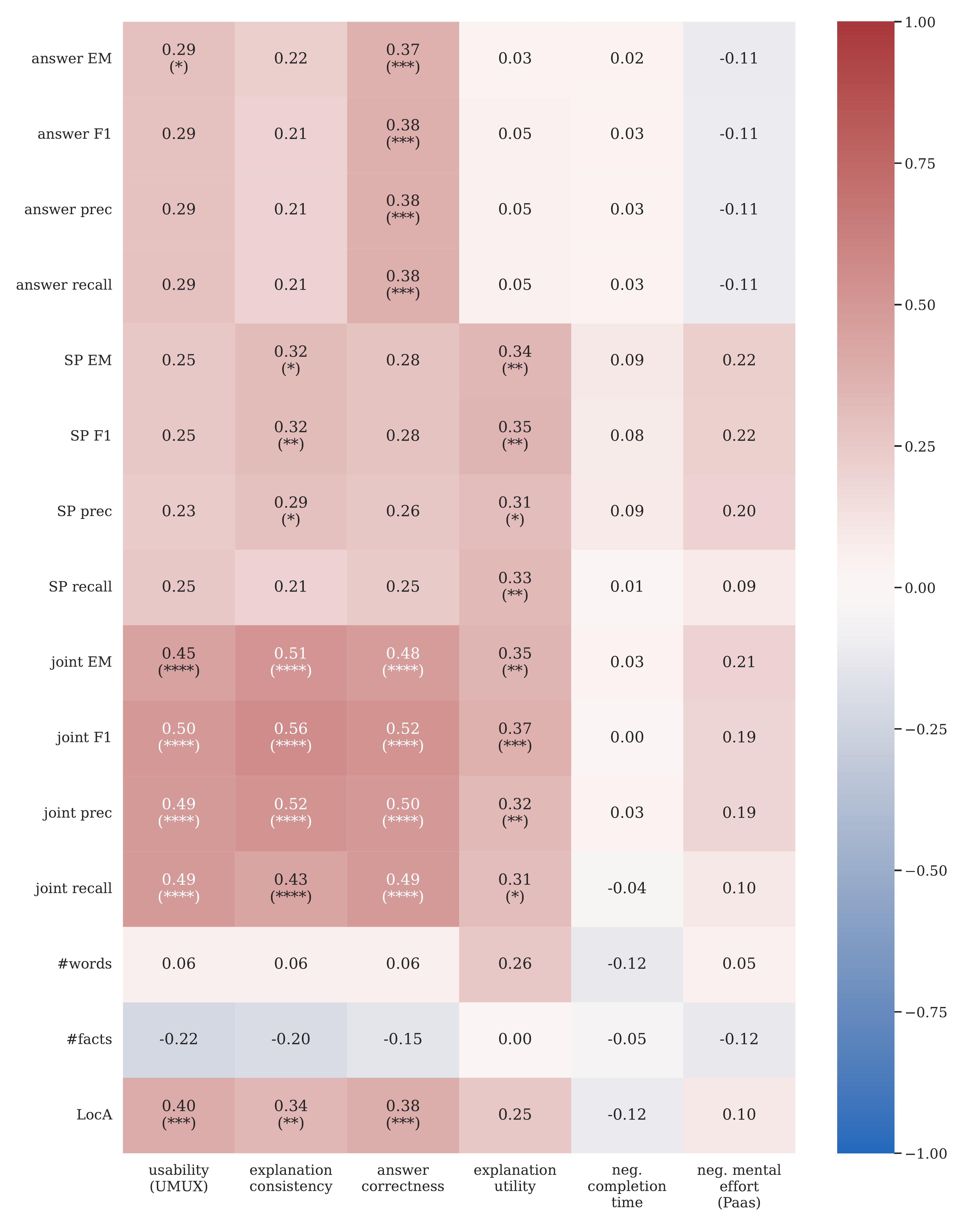}
    \caption{Kendall's $\tau$ correlations (per HIT). Significance levels are corrected using Bonferroni correction. ($\ast$: $p \leq 0.05$, $\ast\ast$: $p \leq 0.01$, $\ast\ast\ast$: $p \leq 0.001$ and $\ast\ast\ast\ast$: $p \leq 0.0001)$}\label{fig:human_machine_correlations_kendall}
\end{figure*}

\Cref{fig:human_machine_correlations_kendall} displays the Kendall's $\tau$ correlations between proxy scores and human ratings.
We additionally provide Bonferroni-corrected significance levels.

We further evaluate (i) grouped weighted $\kappa$ inter-annotator agreements (IAAs) \cite{cohen1968weighted} as an appropriate IAA measure for ordinal responses and (ii) standard deviations to provide an additional perspective on the ratings' variances.
We observe $\kappa=0.42$ / SD$=0.43$ for correctness, $\kappa=0.3$ / SD$=1.88$ for utility and $\kappa=0.33$ / SD$=2.13$ for consistency.
These IAAs and standard deviations signal a low agreement / high variability which is commonly interpreted to correspond to low-quality \textit{annotations}.\footnote{We note that this interpretation can be challenged and low IAAs are not necessary to collect highly reliable data \citep{beigman-klebanov-beigman-2009-squibs}.}
However, we want to emphasize that the purpose of our study is not (and should not be) to collect clean \textit{annotations} of specific explanation instances but instead to capture the relation between automatic scores and intentionally and potentially noisy \textit{subjective human ratings} as these are the exact ratings that constitute human assessment of explanation quality.

\begin{table*}
    \centering
    \resizebox{\textwidth}{!}{
        \begin{tabular}{L{0.38\textwidth} R{0.15\textwidth} R{0.16\textwidth} R{0.09\textwidth} R{0.16\textwidth} R{0.10\textwidth} R{0.19\textwidth}}
            \toprule
            {} &  \textbf{Usability (UMUX)} &  \textbf{Consistency} &  \textbf{Utility} &  \textbf{Answer Correctness} &  \textbf{Mental Effort} &  \textbf{Completion Time (seconds)} \\
            \midrule
            AMGN                        &   86.7 &                    5.8 &                5.6 &                0.9 &                 5.8 &                        80.2 \\
            DecompRC                    &   78.3 &                    5.0 &                4.8 &                0.9 &                 5.8 &                        43.2 \\
            FE2H on ALBERT              &   \textbf{97.5} &           \textbf{6.3} &       6.2 &                1.9 &                 \textbf{4.0} &               81.8 \\
            $\ast$ \textit{gold}                        &   83.3 &                    6.1 &                6.2 &                \textbf{2.0} &        5.6 &                        \textbf{41.4} \\
            $\ast$ \textit{gold-answers-all-facts}      &   85.8 &                    5.0 &                5.6 &                1.8 &                 5.8 &                        75.4 \\
            $\ast$ \textit{gold-answers-random-facts}   &   15.8 &                    2.3 &                2.4 &                1.7 &                 7.8 &                        43.8 \\
            GRN                         &   68.3 &                    5.4 &                5.8 &                1.7 &                 4.8 &                        75.1 \\
            HGN                         &   90.0 &                    \textbf{6.3} &       \textbf{6.3} &       1.9 &                 4.2 &                        64.4 \\
            IRC                         &   83.3 &                    6.0 &                \textbf{6.3} &       1.8 &                 5.8 &                       118.0 \\
            Longformer                  &   86.7 &                    5.9 &                \textbf{6.3} &       1.9 &                 5.0 &                        42.0 \\
            $\ast$ \textit{random-answers-gold-facts}   &   20.8 &                    2.1 &                5.4 &                1.0 &                 4.6 &                        44.4 \\
            $\ast$ \textit{random-answers-random-facts} &   23.3 &                    2.4 &                2.9 &                1.0 &                 5.2 &                        48.7 \\
            S2G-large                   &   88.3 &                    6.1 &                6.1 &                1.8 &                 \textbf{4.0} &               50.9 \\
            SAE                         &   86.7 &                    5.9 &                \textbf{6.3} &       1.8 &                 4.2 &                        86.6 \\
            Text-CAN                    &   86.7 &                    6.0 &                \textbf{6.3} &       1.9 &                 4.6 &                        94.2 \\
            \bottomrule
        \end{tabular}
    }\vspace*{0.1cm}%
    \caption{Human ratings of the systems we assessed within our human evaluation (synthetic systems are marked with ``$\ast$'' and \textit{italics}). Best values are marked \textbf{bold}. Answer correctness ratings are scaled to $[0,2]$ to allow a finer-grained differentiation between systems.
    }
    \label{tab:human_ratings}
\end{table*}

\begin{figure*}
    \centering
    \includegraphics[width=\linewidth]{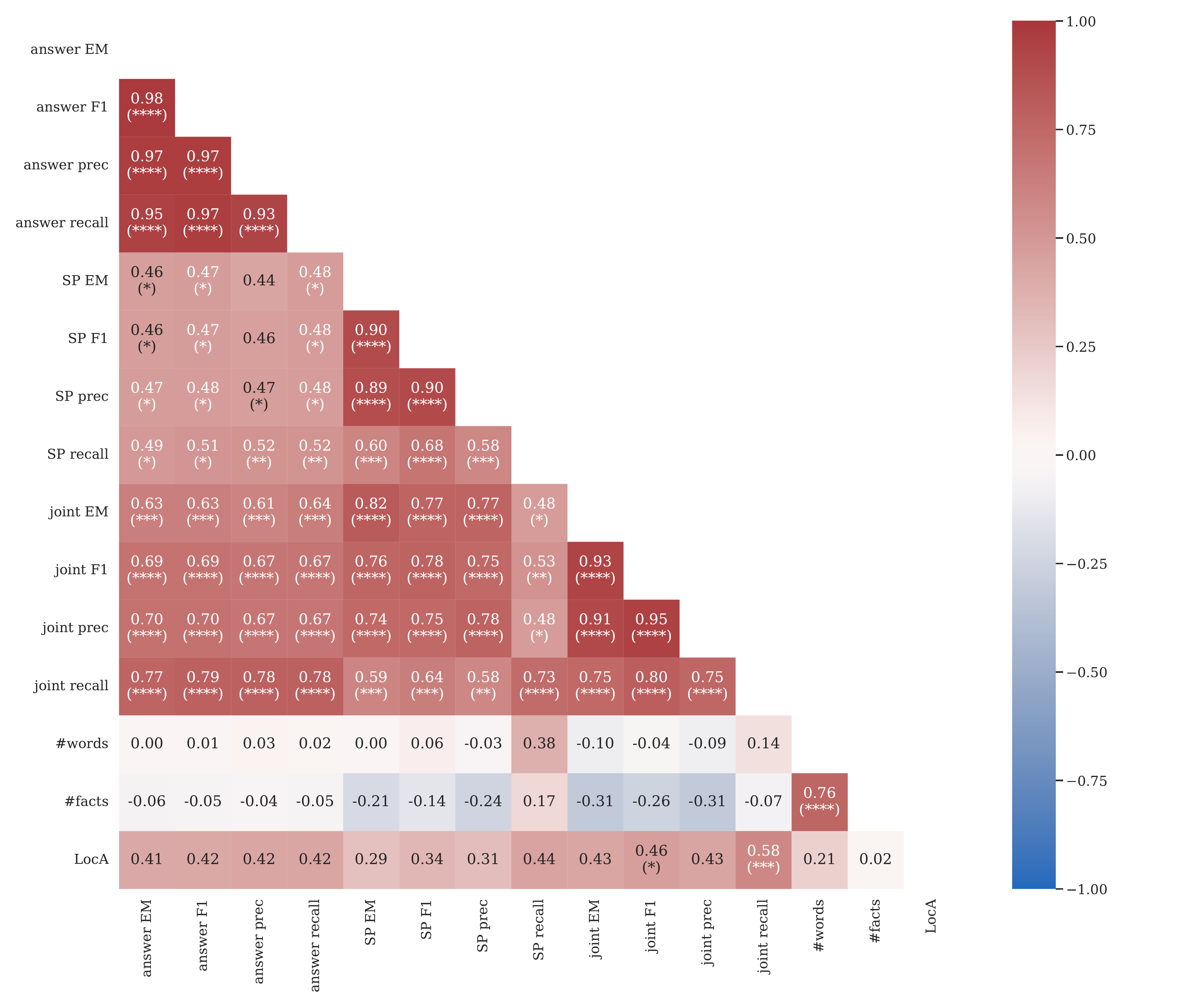}
    \caption{Kendall's $\tau$ correlation coefficients between automatic scores to quantifying model behaviour related to explanation quality on the HotpotQA dataset. Significance levels are corrected using Bonferroni correction. ($\ast$: $p \leq 0.05$, $\ast\ast$: $p \leq 0.01$, $\ast\ast\ast$: $p \leq 0.001$ and $\ast\ast\ast\ast$: $p \leq 0.0001)$}\label{fig:score_score_kendall_correlations}
\end{figure*}

\subsection{Question Pool Size Simulations.}\label{sec:few_questions_simulation}
\begin{figure*}
    \centering
    \includegraphics[width=0.95\textwidth]{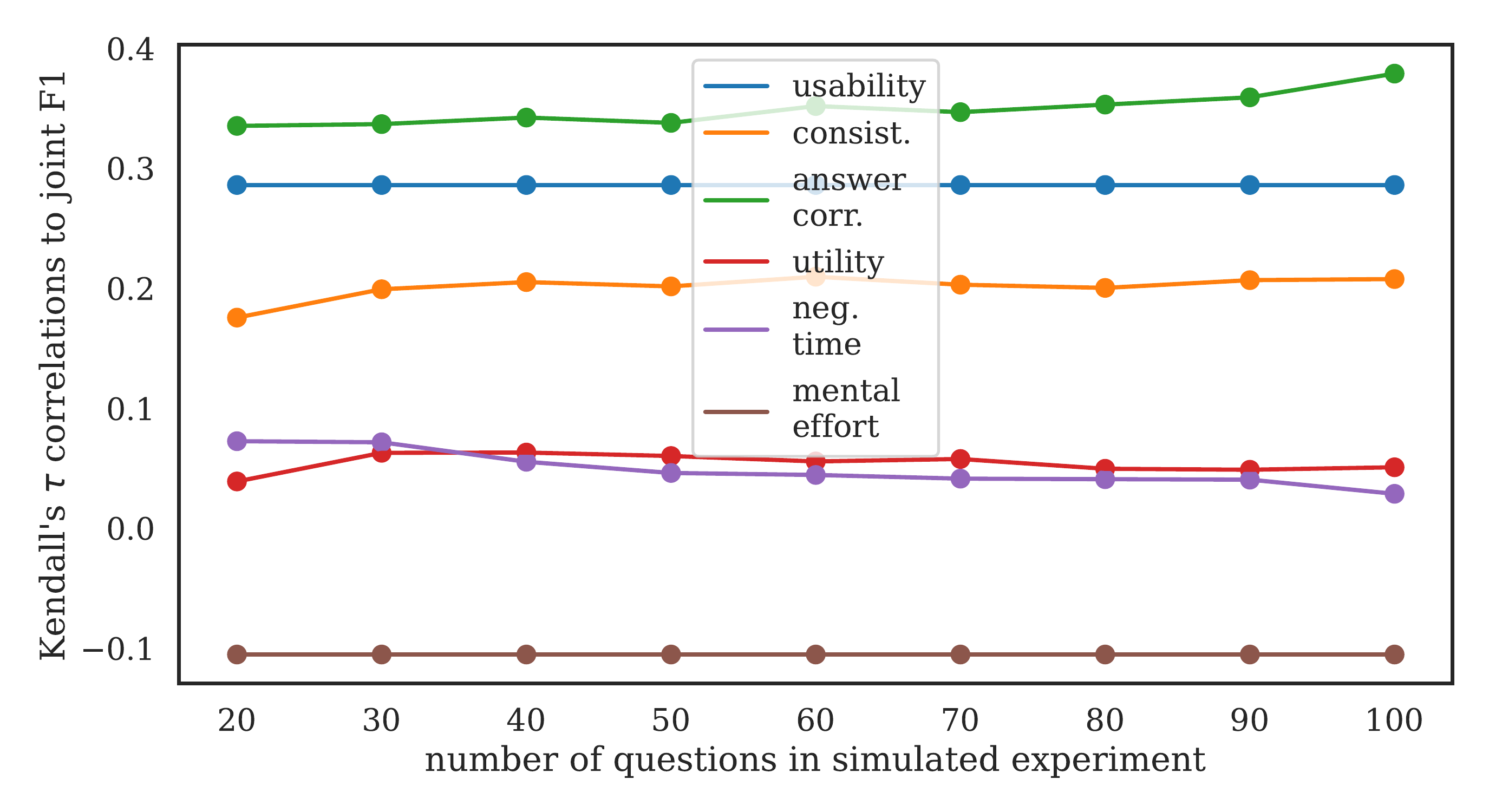}
    \caption{Kendall's $\tau$ correlation coefficients between human ratings and joint-F1.}\label{fig:question_pool_simulation}
\end{figure*}

In order to support our assumption that our pool of 100 questions is sufficiently representative, we simulate
experiments with various question subsets.
\Cref{fig:question_pool_simulation} shows that correlations already stabilize for 20 questions and that there are no qualitative or
quantitative differences to using 100 (all $\tau$ differences<=0.04).

\end{appendices}
\FloatBarrier

\bibliography{anthology,custom}% common bib file

\end{document}